\documentclass[10pt,twocolumn,letterpaper]{article}

\usepackage{iccv}
\usepackage{times}
\usepackage{epsfig}
\usepackage{float}
\usepackage{graphicx}
\usepackage{amsmath}
\usepackage{amssymb}
\usepackage{caption}
\usepackage{subcaption}
\usepackage{adjustbox}
\usepackage{makecell}
\usepackage{bm,multirow}

\long\def\comment#1{} 

\newcommand{\diag}{\operatorname{diag}}

\newcommand{\xmath}[1] {\ensuremath{#1}\xspace}
\newcommand{\blmath}[1] {\xmath{\bm{#1}}}

\newcommand{\W}{\blmath{W}}

\newcommand{\Ab}{{\blmath A}}
\newcommand{\Bb}{{\blmath B}}
\newcommand{\Cb}{{\blmath{C}}}

\newcommand{\Ib}{{\blmath I}}

\newcommand{\Rb}{{\blmath R}}

\newcommand{\Tb}{{\blmath T}}

\newcommand{\Wb}{{\blmath W}}
\newcommand{\Xb}{{\blmath X}}
\newcommand{\Yb}{{\blmath Y}}

\newcommand{\db}{{\blmath d}}

\newcommand{\xb}{{\blmath x}}
\newcommand{\yb}{{\blmath y}}

\newcommand{\Rd}{{\mathbb R}}

\newcommand{\beq}{\begin{equation}}
\newcommand{\eeq}{\end{equation}}
\newcommand{\beqa}{\begin{eqnarray}}
\newcommand{\eeqa}{\end{eqnarray}}

\usepackage[pagebackref=true,breaklinks=true,colorlinks,bookmarks=false]{hyperref}
\iccvfinalcopy 

\def\iccvPaperID{7200} 

\ificcvfinal\fi

\begin{document}

\title{Diagonal Attention and Style-based GAN for Content-Style Disentanglement in Image Generation and Translation}
\author{Gihyun Kwon \quad\quad Jong Chul Ye\\
Korea Advanced Institute of Science \& Technology (KAIST),  Korea\\
\tt\small \{cyclomon,jong.ye\}@kaist.ac.kr
}
\twocolumn[{%
\renewcommand\twocolumn[1][]{#1}%
\maketitle
\begin{center}
\centering
\includegraphics[width=0.9\linewidth]{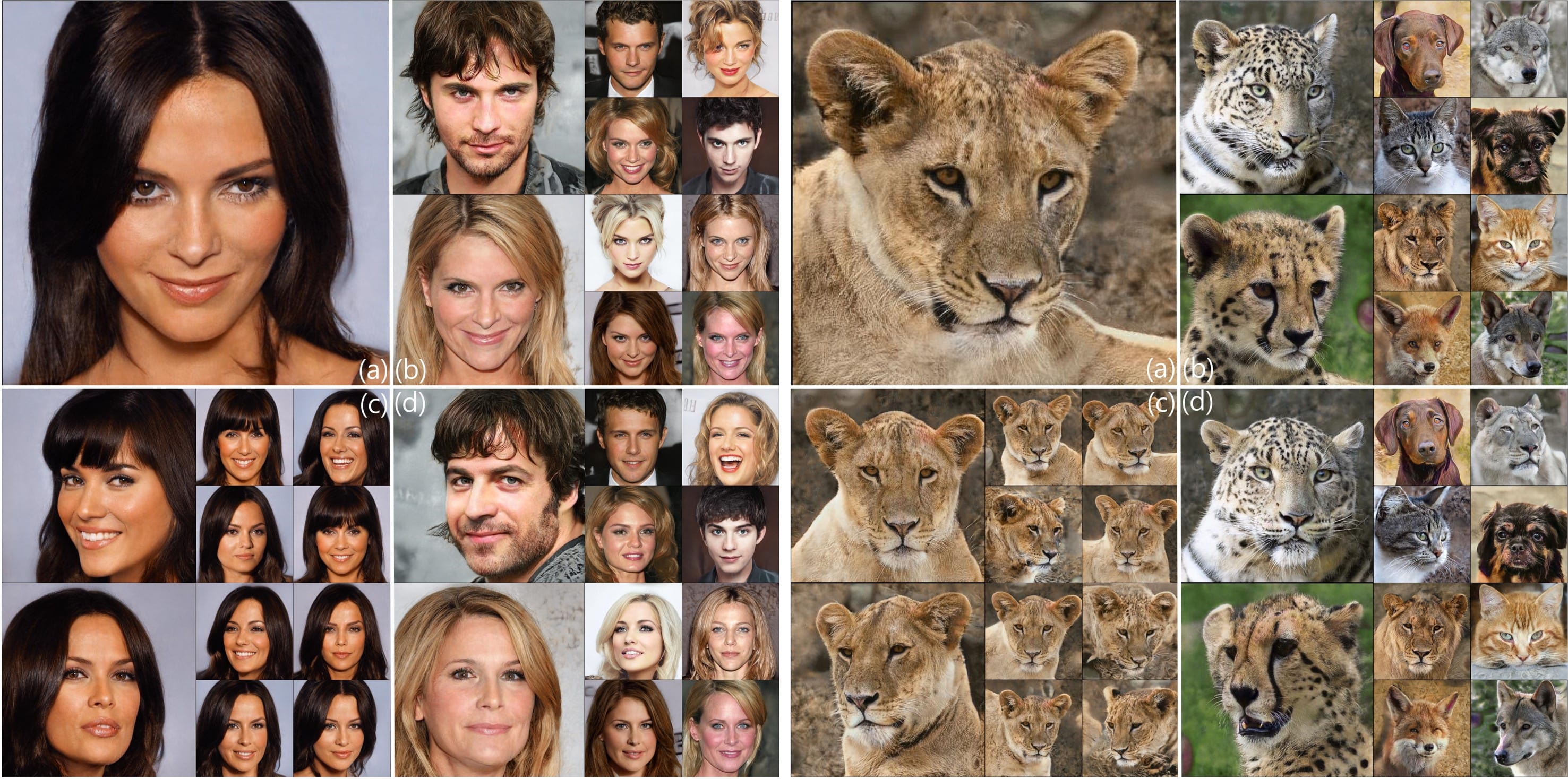}
 \vspace*{-0.2cm}
\captionof{figure}{Full-resolution images synthesized with different diagonal attention (DAT) content codes and AdaIN style codes.  (Left) Generated 1024$\times$1024 images by our method 
trained using CelebA-HQ. (Right) Generated 512$\times$512 images by our method 
trained using AFHQ.  (a) A source image generated from arbitrary content and style codes.  Samples generated by (b) varying style codes with  fixed content codes, (c) varying content codes with  fixed style codes, and  (d) varying both codes. We can see that
content code controls the spatial content  such as directions, whereas style codes affects the styles like gender, hair color, etc.}
\label{fig:first}
\end{center}}]
\begin{abstract}
One of the important research topics in image generative models is  to  
disentangle the  spatial contents and styles for their separate control.
Although StyleGAN can generate  content feature vectors from random noises,  the resulting spatial content control is primarily intended for minor spatial variations,  and the disentanglement of global  content and styles is by no means complete.
Inspired by a mathematical understanding of normalization and attention, 
here we present a novel hierarchical adaptive Diagonal spatial ATtention (DAT) layers to separately manipulate the spatial contents from styles in a hierarchical manner.
Using DAT and AdaIN, our method 
enables coarse-to-fine level disentanglement of spatial contents and styles. 
In addition, our generator can be easily integrated into the GAN inversion framework so that the content and style of translated images from multi-domain image translation tasks can be flexibly controlled.
 By using various datasets, 
 we  confirm that the proposed method not only outperforms the
existing models in disentanglement scores, but also provides more flexible control over spatial features in the generated images. 

\end{abstract}

\section{Introduction}

Recent development of Generative Adversarial Networks (GAN) \cite{gan} has enabled the generation of high-quality images that are  indistinguishable to the human eye. Despite its high performance, 
 disentangling the attributes of the generated images is still an open problem.

\begin{figure*}[!t]
\centering
  \includegraphics[width=0.9\linewidth]{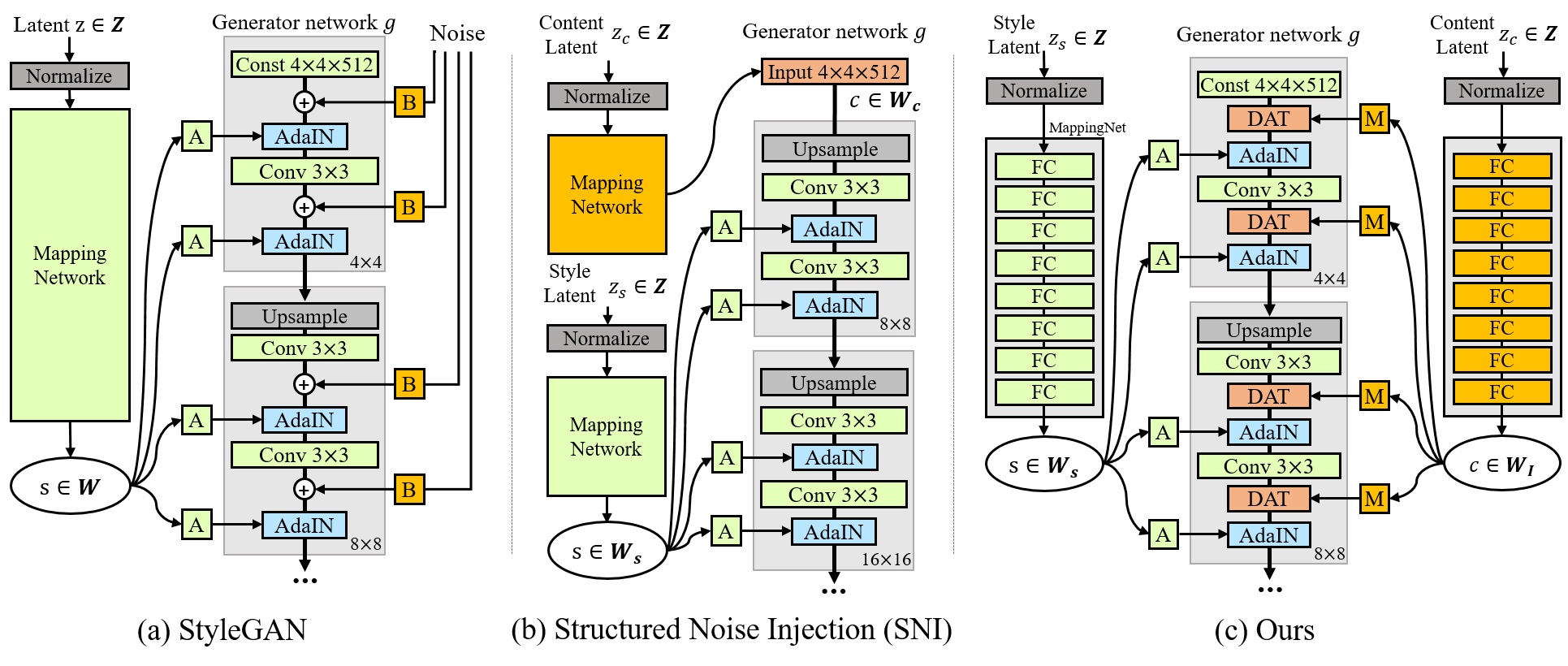}
\caption{Various style and content disentanglements: (a) StyleGAN with style and content control by AdaIN and per-pixel noises, respectively. 
(b) Structured Noise Injection (SNI) with additional content codes as an input tensor for a generator network. 
(c) Our approach with diagonal attention (DAT) and AdaIN  for the  content and style disentanglement. }
\label{fig:method}
\end{figure*}

For example, the content and style disentanglement is an important issue in many image generation tasks such as faces.
Here, contents refer to the spatial information such as face direction, expression, whereas
styles are related with other features such as color, makeup, gender.
StyleGAN \cite{stylegan}, which shows the state-of-the-art performance in image generation, tries to disentangle the style and content
using the AdaIN codes \cite{adain} and the content feature vectors from random per-pixel noises, respectively.
 The AdaIN layer then combines the style  and the content features  to generate more realistic features at each resolution
 (see Fig.~\ref{fig:method}(a)). 
 However, the content control by per-pixel noises is mostly for minor spatial variations so that the disentanglement of global  contents and styles is by no means complete.

Recently, generative models that simultaneously use AdaIN and independent content latent codes \cite{scgan,sni} have shown good performance in separating global  style and content information.
For example, in recent structured noise injection (SNI) approach \cite{sni},  the latent code for content  is generated by an additional neural network, which is used as an input tensor of the image generator composed of subsequent layers for style control using AdaIN (see Fig.~\ref{fig:method}(b)). Although SNI showed good performance in disentanglement,
one of the major drawbacks  is that the size of the input tensor is limited to relatively small resolution (e.g. 4$\times$4). 
Therefore, the intended content control often fails to work properly due to the limited capacity. 

To address these issues, here we introduce a novel {Diagonal spatial ATtention (DAT)} module to manipulate the content feature in a hierarchical manner. Specifically, the content code is applied to multiple layer features as diagonal attention maps at various resolutions as shown in Fig.~\ref{fig:method}(c). Despite the simplicity of diagonal attention,
one of the
important advantages of DAT is that the image content and style 
can be modulated independently in a symmetric manner; and
 similar to AdaIN for the styles,  DAT enables the hierarchical control of the spatial content.
These lead to an
effective disentanglement of the content and style components in generated images

In addition, our method can be easily integrated into  the state-of-the-art GAN inversion \cite{idinvert}, allowing much more flexible post-hoc control of the content and style in the translated images from the multi-domain image translation.

\section{Related works}
\subsection{Spatial attention}
Spatial attention helps the visual learning tasks by highlighting the specific regions that contain important information.
Several methods have used spatial attention to improve performance on some visual tasks: object detection \cite{vsgnet,csanet}, semantic segmentation \cite{dual,cross}, image captioning \cite{7sca}, and so on.  
Spatial attention has been further extended to non-traditional image generation tasks. For example, self-attention GAN \cite{sagan} 
enhances the generation of geometrical and structural patterns. 
In the image-to-image translation tasks, recent methods achieved realistic generation performance with attention maps that focus on spatial areas of targeted objects \cite{mejjati,spagan} or face components \cite{face,attgan,semanticface}.


\subsection{Disentangled representation}

For disentangled image generation, several approaches have been proposed. Direct approaches rely on increasing the connection between the latent and image spaces \cite{infogan,infogancr},  a specialized training to constrain the latent space \cite{unified,semantic},  manipulating the prior distribution of latent \cite{disentangling}, or using external attribute information \cite{explicit}. 
Other approaches for disentanglement rely on  the hierarchical structure of networks using layer-dependent latent variables in VAE to encode the disentangled attributes \cite{lvae,vlae,progVAE},  using a tree-like latent variable structure \cite{tree}, or synthesizing image components in several stages \cite{finegan}. 
Despite the theoretical motivations, the above methods often suffer from poor generation quality due to limited network capacity or from disadvantages due to the need for additional attribute labels.

Recently, several authors propose to use an additional latent vector which controls independent attributes from the original one. For example, SC-GAN \cite{scgan} separates style and content information using AdaIN along with input content codes. 
Additionally, there are methods which employ style-content disentanglement to improve the style transfer \cite{artsty}, and image translation \cite{improving}. Recently, a state-of-the-art style-content disentanglement was proposed in \cite{sni}, which allows to control various spatial attributes by injecting structured noise as an input tensor of StyleGAN. 
Specifically, as shown in Fig.~\ref{fig:method}(b), a content latent $z_c$ is processed by a specific neural network and directly used as an input tensor of the generator network. 
However, one of the drawbacks of this approach is its asymmetric architecture: although the 
style can be manipulated in a multi-resolution manner using  hierarchical AdaIN layers, the content is controlled using a single input.

\subsection{Our contributions}

The architecture of our method, which we call the diagonal GAN, is shown in Fig.~\ref{fig:method}(c) and has several
advantages over existing disentanglement methods.
\begin{itemize}
\item
In contrast to the original styleGAN  in Fig.~\ref{fig:method}(a), the content and style code generation
is symmetric by using similar code generators. 
Similar to the AdaIN layer, diagonal attention layer (DAT) enables the 
spatial control of the content in a hierarchical way that 
is difficult by SNI in Fig.~\ref{fig:method}(b).
\item Although existing attention approach is implemented by multiplying a fully populated attention matrix,
  our approach is unique in that it uses a diagonal attention matrix to manipulate content information. While using a simple network architecture, this is a more efficient method as it enables much more powerful control of global content compared to the baseline StyleGAN model (see Fig. 2(a)).
\end{itemize}

\section{Theory}


\subsection{Mathematics of Normalization and Attention}

To understand the motivation of the proposed DAT layer, here we provide some mathematical
analysis of existing normalization and attention modules in neural networks. Our analysis shows that
 the normalization and spatial attention have similar structure that can be exploited for style and content  disentanglement.


Specifically, let $H,W$ and $C$ denote the height and width of the feature map, and the number of the feature channels, respectively.
Then, for a given feature map $\Xb\in \Rd^{HW\times C}$, the AdaIN normalization layer output $\Yb\in \Rd^{HW\times C}$ can be represented as
follows:
\begin{align}\label{eq:T}
\Yb = \Xb\Tb +\Rb
\end{align}
where 
the channel-directional transform $\Tb$ and the bias $\Rb$ are learned from the statistics of the feature maps.
Specifically, $\Tb$ is a {diagonal} matrix that is calculated  as the ratio of the standard deviations of the channel-wise input  and target features,
and $\Rb$ is the bias term that converts the input mean to the target mean. 

Similarly, the spatial attention can be represented by
\begin{align}\label{eq:A}
\Yb = \Ab\Xb
\end{align}
where $\Ab\in \Rd^{HW\times HW}$ is a fully populated matrix that is calculated from its own feature for the case of self-attention, or with the help of other domain features
for the case of cross-domain attention.  
Since the transformation matrix $\Ab$ is applied to pixel-wise direction to manipulate the feature values of a specific location, it can control the spatial information such as shape and location. 

In  styleGAN,  the content code $\Cb$, which is generated from per-pixel noises via the scaling network $\Bb$ (see Fig.~\ref{fig:method}(a)), is added to the feature $\Xb$ before the AdaIN layer. This leads to the following feature transform:
\begin{align}
\Yb = (\Xb+\Cb)\Tb +\Rb=  \Xb\Tb + {\Rb+\Cb\Tb}
\end{align}
Accordingly, the last term, $\Cb\Tb$, works as an additional bias term,  which is different from
the spatial attention  that  is multiplicative to the feature $\Xb$ (see \eqref{eq:A}). Although one could potentially generate
$\Cb$ so that the net effect is similar to $\Ab\Xb$, this would require a complicated content code generation
network. This explains the fundamental limitation of the content control in the original  styleGAN.

\subsection{Diagonal Attention (DAT)}

If   normalization and attention in \eqref{eq:T} and \eqref{eq:A} are applied together, 
the output feature
can be represented by
\begin{align}\label{eq:key}
\Yb = \Ab\Xb\Tb+\Rb
\end{align}
One of the most important observations in this paper
is that the combined equation \eqref{eq:key} is the key for systematic
style-content disentanglement.
Specifically,  $\Tb$  in \eqref{eq:key} from AdaIN  layer  is a diagonal  matrix obtained from a style code generator.
Mathematically, the diagonal  matrix $\Tb$
control the {\em row space}  of the feature  $\Xb$, which turns out to be the style control.
Accordingly, we conjecture that the spatial content can be controlled by manipulating
the remaining  factor: the {\em column space} of the feature $\Xb$. Mathematically,
this can be easily implemented by \eqref{eq:key} using a diagonal attention matrix $\Ab$ that is obtained from another content code generator.
%
The diagonal attention and diagonal normalization are then complimentary
to each other, which are applied to different axes of the feature tensor to simultaneously control the two independent factors of the feature tensor $\Xb$.
Furthermore, due to the symmetric role of AdaIN and DAT,
they can  be applied at each layer in a hierarchical manner
 as shown in Fig.~\ref{fig:method}(c).

\begin{figure}[!t]
    \includegraphics[width=0.9\linewidth]{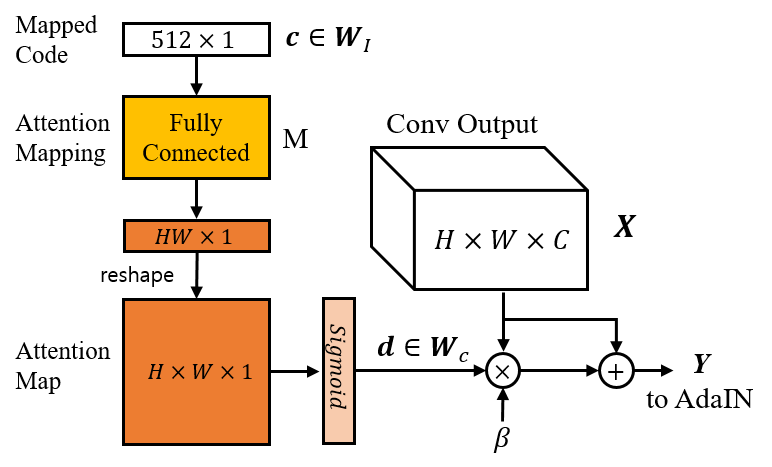}
    \caption{Generation of differential attention. With the attention mapping network $M$, the content code $c$ is converted into an attention map. The map is multiplied element-wise to  the convolution features. {Since the attention maps of each layer contribute to the content information, they work as independent codes in the space $\Wb_c$.}}
    \label{fig:att_map}
\end{figure}
\begin{figure*}[!t]
\centering
\includegraphics[width=0.95\linewidth]{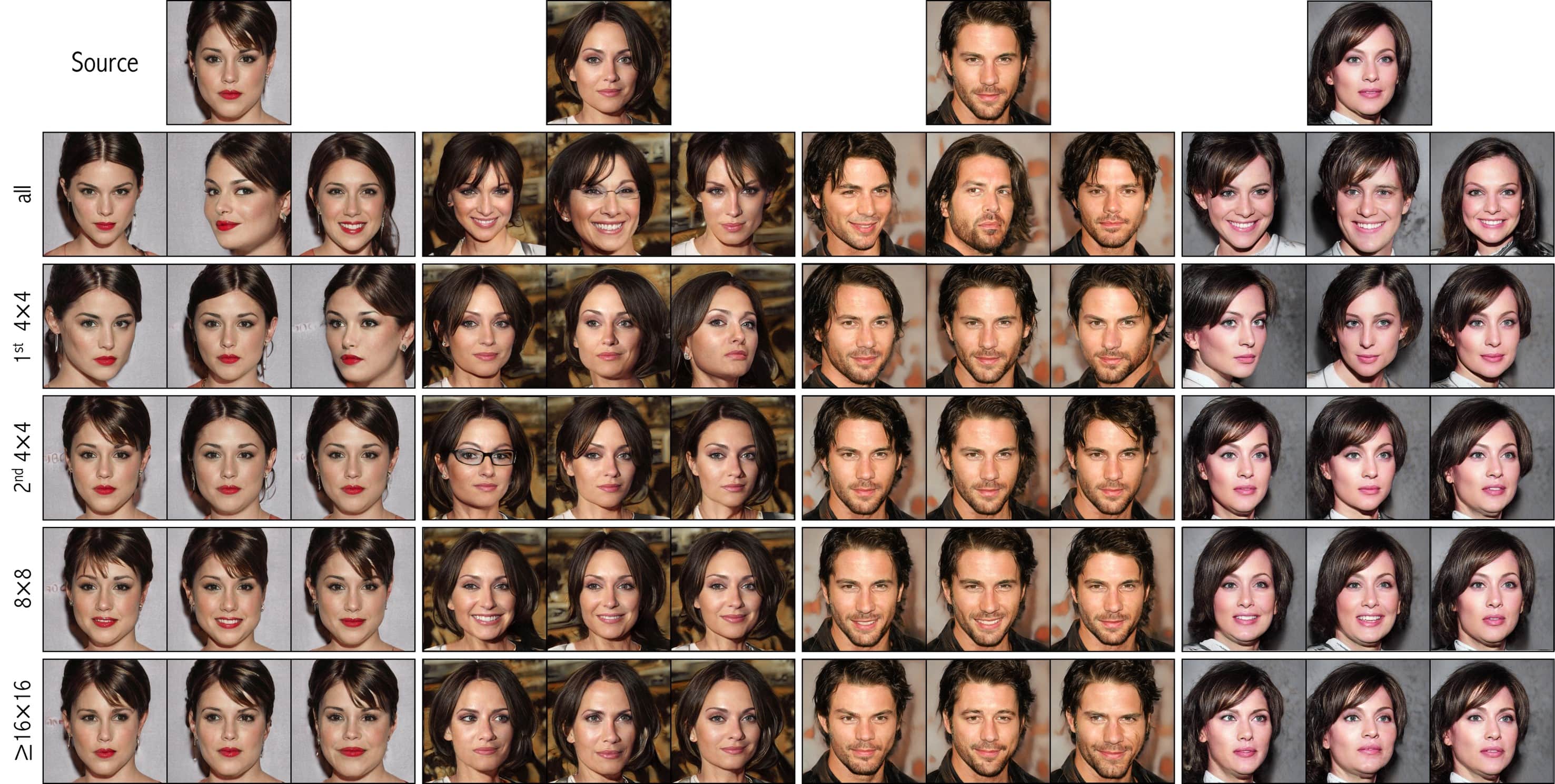}
\vspace*{-0.3cm}
\caption{Our  1024$\times$1024 full-resolution examples with fixed style and different content codes. The source images on top row are sampled from arbitrary DAT content and AdaIN style codes. The images on the second row are generated with changing content codes at entire layers  under the fixed style code.  
The images in the following rows are sampled with changing the content codes at specific layers while fixing those of other layers. 
 The  hierarchical DAT layer can selectively control the extent of attribute changes.
}
\label{fig:layerwise}
\end{figure*}

%

Specifically, Fig.~\ref{fig:method}(c) describes the overall architecture of our proposed model. We adopt a method of using two different latent codes. In addition to the style latent code $z_s$, we use an independent latent code $z_c$ to control the content information. 
More specifically, our style code $z_s$ is mapped into a linearly distributed space $\Wb_s$ by several MLPs. Then the mapped code $s$ is transformed into the parameters that can be applied to multiple layers as AdaIN. 
Similar to the style code mapping, the content code $z_c$ is also mapped to a linear space $\Wb_I$ through a mapping function consisting of a series of MLPs. The mapped intermediate content code $c$ can change the spatial information of the convolution output through the proposed attention mapping.

Figure \ref{fig:att_map} is a detailed diagram of our attention mapping network.
Here, rather than directly estimating the diagonal component of $\Ab$, we are interested in estimating the perturbation from the identity attention.
Specifically,  the mapped content code $c$ is converted into a vector with $HW\times1$ dimension. Then the vector is reshaped into a differential attention map $\db \in \Wb_c$ which has the same spatial dimension $H\times W$ to that of the output from convolution layer. In order to avoid the undesired artifacts from excessive diversity in the attention map, we limit the value range of the differential attention with the help of  {{sigmoid}} activation. Thanks to the diagonal attention map,
the network output  
 is then element-wise multiplied with feature map  at each channel, which  is added to the original feature map.
 In this stage, we use an additional parameter $\beta$, allowing the 
attention map of the network to learn the layer-wise contribution of content control.  Since the contribution of attention can be calibrated  by $\beta$ depending on whether the layer is responsible for  minor or major changes, an artifact from overemphasizing minor changes can be prevented. 

Accordingly, the resulting feature output can be represented by
\begin{align}
\yb_i =  \xb_i+\beta\db\odot \xb_i = \left(\Ib+\beta\diag(\db)\right)\xb_i
\end{align}
where $\diag(\db)$ denotes the diagonal matrix whose diagonal elements is $\db$.
This suggests that the resulting diagonal attention matrix is 
$$\Ab=\left(\Ib+\beta\diag(\db)\right) \ .$$

The DAT layer can also easily incorporate the per-pixel noises used in StyleGAN. 
However, care should be taken as
 per-pixel noise is only additive so that it can change minor spatial variations, whereas
our diagonal spatial attention is multiplicative so that we can control  global spatial  variations.

\section{Method}
\subsection{Loss function}
Our implementation is inspired by the original StyleGAN paper and the source code\footnote{https://github.com/rosinality/style-based-gan-pytorch} implemented on PyTorch.   Similar to StyleGAN, we choose the non-saturating loss with {\it{R$_1$}} regularization for adversarial loss \cite{mescheder2018training}. 
We also use Diversity-Sensitive (DS) loss \cite{ds} to encourage the attention network to yield diverse maps. 
More specifically, if we sample two content codes $z_c^1$, $z_c^2$ and a style code $z_s$, our DS loss is defined as:
\begin{equation}
    L_{ds} = \max\left(\lambda - \|G(z_s,z_c^1) - G(z_s,z_c^2)\|_1,0\right)
\end{equation}
where $G(z_s,z_c)$ is our generator with respect to the style code $z_s$ and the content coder $z_c$, respectively.
The objective of our DS loss is to maximize the $L_1$ distance between the generated images from different content codes with same style. However, directly optimizing the negative $L_1$ loss will lead to the explosion of the loss value. Therefore, we penalize DS loss with threshold $\lambda$ so that the distance will not exceed $\lambda$. Accordingly, our total loss function is described as: 
\begin{equation}
 L_{total} = L_{adv} + L_{ds}   \ ,
\end{equation}
where $L_{adv}$ is the adversarial loss.





\subsection{Experiment Settings}


\begin{figure*}[t]
\vspace*{-.7cm}
\centering
\includegraphics[width=0.9\linewidth]{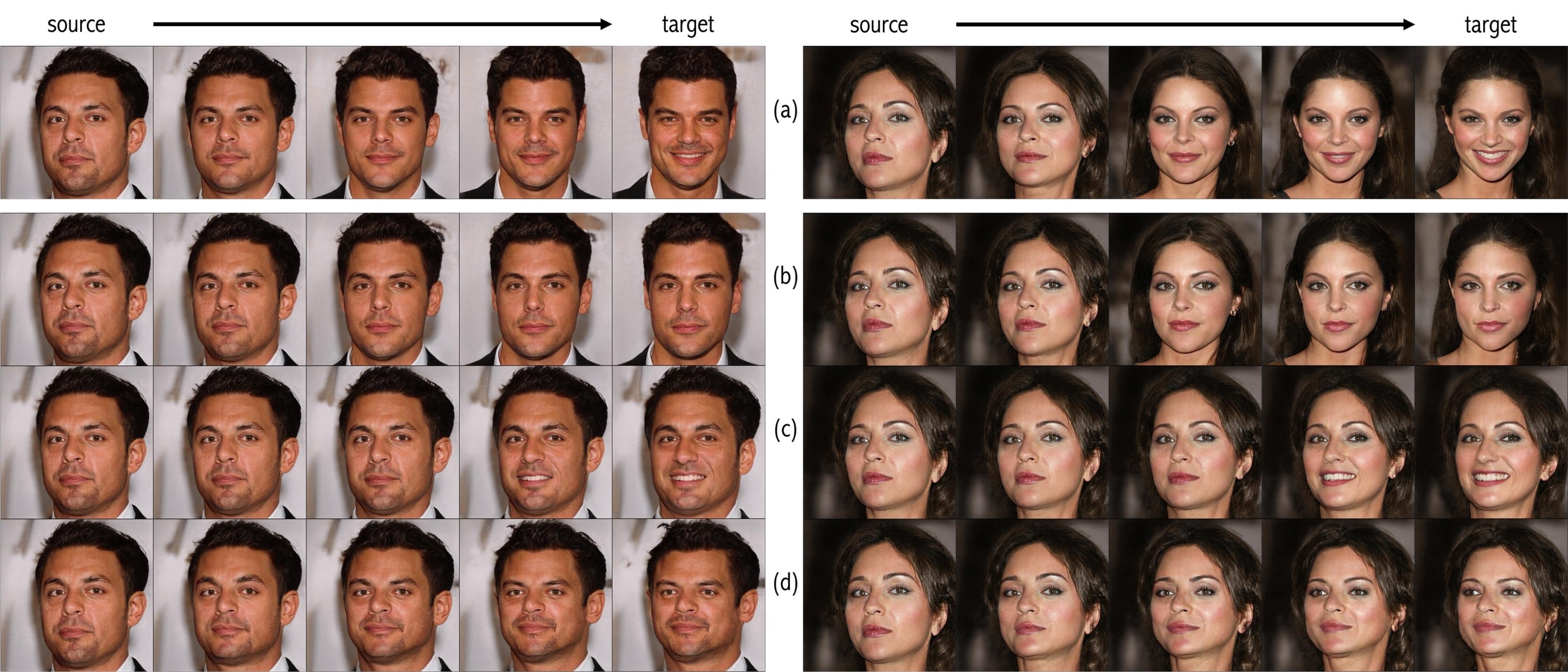}
\vspace*{-.2cm}
\caption{CelebA-HQ results from interpolated content codes. The images at each column are sampled from the same (interpolated) content code. (a) Results with interpolating content codes of all layers. (b) Interpolating 4$\times$4 layer content codes. (c) Interpolating 8$\times$8 layer content codes. (d) Interpolating content codes after 16$\times$16 layer.
}
\vspace*{-.3cm}
\label{fig:intp}
\end{figure*}

For qualitative evaluation, we report the results from the model trained on 1024$\times$1024 CelebA-HQ \cite{progan} and 512$\times$512 AFHQ \cite{starganv2}. 
 In Supplementary Material, we also provide experimental results using  flowers \cite{flower}, birds \cite{bird}, cars \cite{car} data sets.
Considering the number of parameters for attention mapping at high resolutions, we include the DAT layers up to the resolution of 256$\times$256. 
For quantitative evaluation, we compared our method with the baseline models that use input noises for content control. Among several methods using this approach, we use the state-of-the-art SNI \cite{sni} as a representative method. For fair comparison, we also included SNI trained with content DS loss as a baseline.
We also use original StyleGAN results with per-pixel noises as another comparative model. 
For comparative studies with various parameter settings, we trained the models at the reduced resolution of 256$\times$256 using  500K iterations (total of $\sim$4.7M samples).  
As baseline SNI presented results on models with and without adding per-pixel noises, we showed our results on both conditions.
When training our models, we set the parameter $\lambda$ in the DS loss as 0.3, as it showed the best performance. For more experimental settings, see Supplementary Materials.

For quantitative metrics, we use FID \cite{fid} for measuring the image quality and Perceptual Path Length (PPL) for measuring the disentanglement. PPL was first proposed in StyleGAN \cite{stylegan} to measure the perceptual distance between output images obtained with slightly changing the interpolated codes. A low PPL value means better disentanglement, since there is little interference of irrelevant features between two latent points. This can be also interpreted to mean that the latent space follows the linear trend. {To measure the performance of the mapping networks that map both style and content code into their respective linear spaces, we compare the disentanglement performance by the PPL in the $ W $ (i.e. $W_s$ {and} $W_c$) space.}

\section{Experimental Results}
\subsection{Qualitative Evaluation}
\noindent\textbf{Content and Style Disentanglement:}
Fig.~\ref{fig:first} illustrates full-resolution images synthesized with different DAT and AdaIN codes. 
The left panel shows generated 1024$\times$1024 images by our method 
trained using CelebA-HQ, whereas the right panel shows  generated 512$\times$512 images by our method 
trained using AFHQ.  
For a given source image (a), which is generated from arbitrary DAT content and AdaIN style codes,
the images in (b) show the generated samples with varying style codes and the content code, whereas
(c) illustrates samples with varying content codes and the fixed style codes.  We can clearly see the effect of content code:
the content of faces, such as the direction and components, vary. This is different from the effect of style codes in (b), which changes the hair color, gender, etc., while the face direction and components are fixed. By using specific style and content codes in (b)(c), the images in (c)  shows that the face direction and components follow the content in (b), whereas
 the hair color, gender, etc are controlled by the style in (b).  This experiments clearly indicates the powerful content and style disentanglement by our method.

\noindent\textbf{Hierarchical Content Disentanglement: }
We also show the hierarchical disentanglement ability by controlling diagonal attention map at each layer. The generated samples are presented in Figure \ref{fig:layerwise}. 
%
%
 The source images on top row are sampled from arbitrary DAT content and AdaIN style codes. The images on the second row are generated with changing entire content code under the fixed style codes. We can observe the variations of entire spatial attributes including shape, rotation and facial expressions with consistent styles. The images in the following rows are sampled with changing the content codes at specific layers while fixing those of other layers. 
The first DAT at 4$\times$4 layer mainly focuses on geometrical change, and the second 4$\times$4 DAT changes hairstyles and eyes accessories. 
The  8$\times$8 DAT layer mainly changes the lower part of facial expressions, and the DAT layers at higher resolutions give relatively minor variations such as hair curls and eyes. 
 
Quantitatively, our CelebA-HQ model showed satisfying performance of {7.32} in FID, compared with 5.17 of original StyleGAN.

\begin{figure}[!t]
\centering
\includegraphics[width=0.90\linewidth]{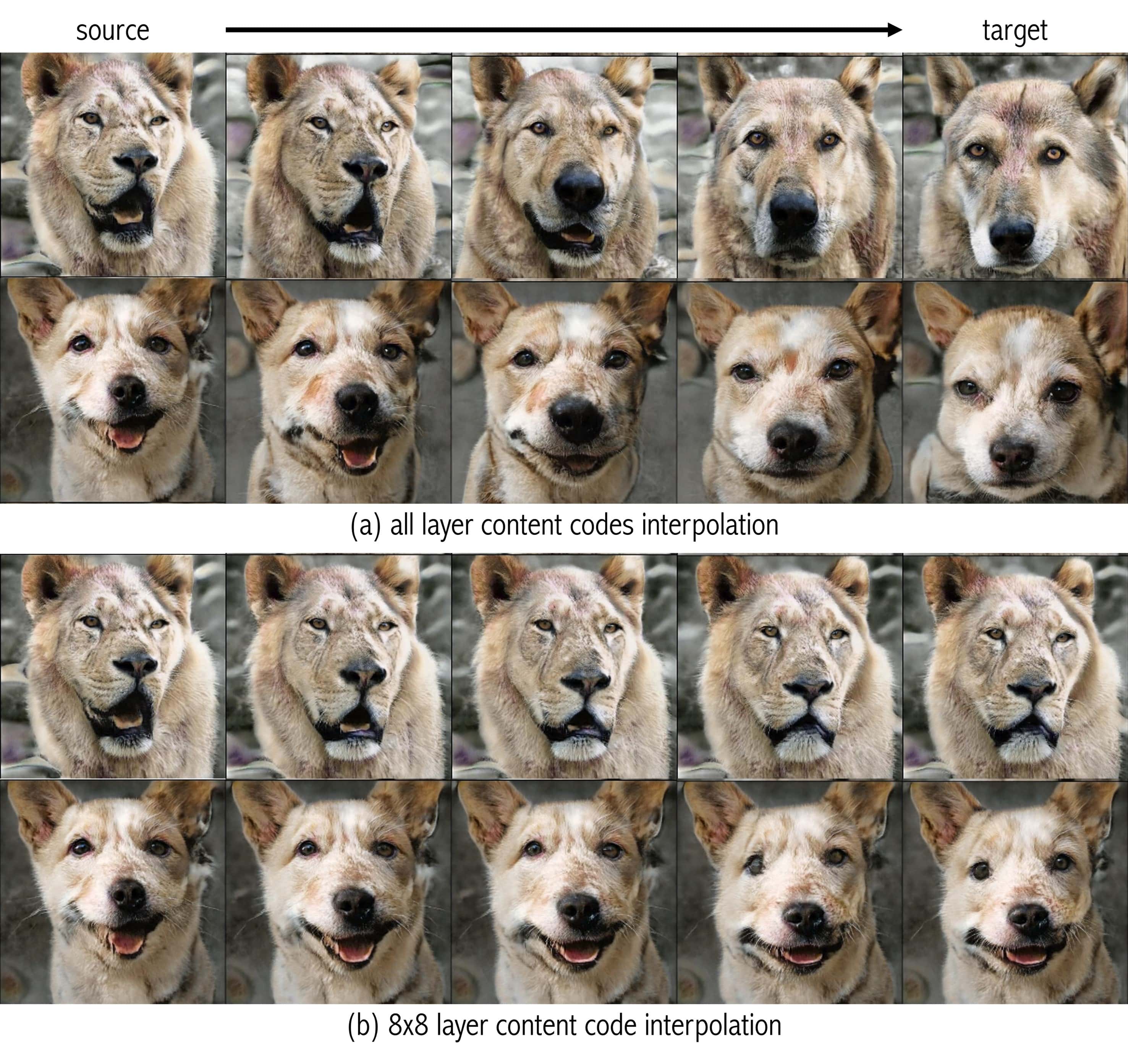}
\vspace*{-.3cm}
\caption{AFHQ results for content code interpolation. (a) Results with interpolating content codes of all layers. (b) Interpolating 8$\times$8 layer content codes.
}
\vspace*{-.3cm}
\label{fig:af_intp}
\end{figure}

\noindent\textbf{Hierarchical Latent Interpolation: }
Figure \ref{fig:intp} show the generated examples by interpolating DAT content codes $c\in W_I$ between two randomly sampled points with fixed style. 
The first row shows results from interpolating content codes of all layers, whereas the rest of the rows illustrate the results by interpolating specific layer content codes.
Although  similar latent interpolation in the first row (Figure \ref{fig:intp}(a)) could be done by StyleGAN, the fine spatial detail
interpolation in Figure \ref{fig:intp}(b)-(d), such as mouth expressions, is not possible in StyleGAN.
On the other hand, our method allows hierarchical content interpolation by interpolating the specific layer content codes. {This can be also seen in our AFHQ results in Figure \ref{fig:af_intp}. In addition to changes of global content in Figure \ref{fig:af_intp}(a), we can smooth change the specific attribute of mouth by  a specific layer content code interpolation as shown in Figure \ref{fig:af_intp}(b). For additional interpolation results using other data sets, see Supplementary Materials.}
%
%
%

\begin{figure}[!t]
\centering
  \includegraphics[width=0.8\linewidth]{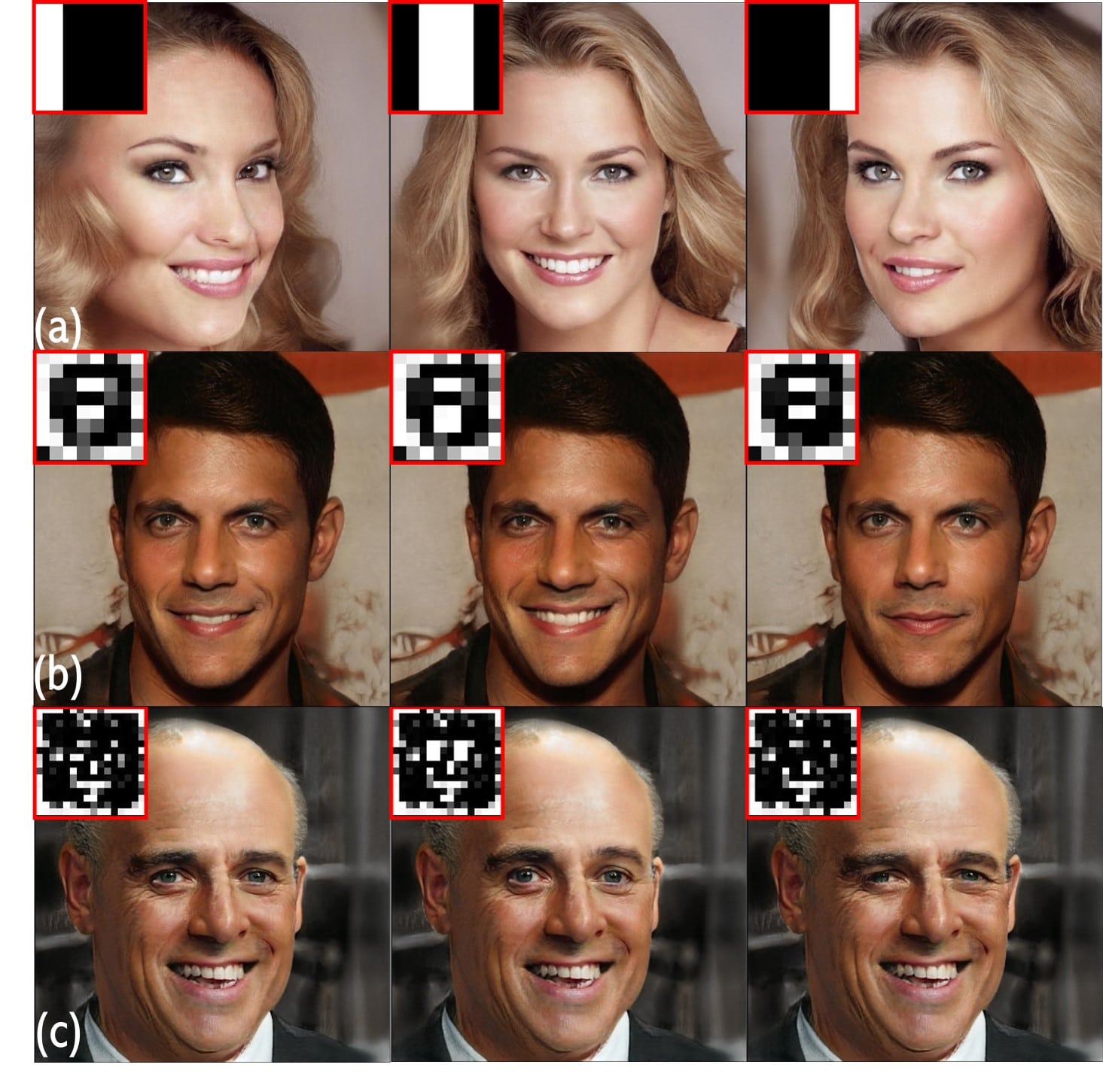}
 \vspace*{-.2cm}
\caption{Direct attention map manipulation. By controlling the specific areas of attention, we can selectively change the facial attributes.
Results from changing (a) the first 4$\times$4 attention map, (b) the 2nd 8$\times$8 attention map,
and (c) the second 16$\times$16 attention map.}
\label{fig:map}
\end{figure}

\noindent \textbf{Direct Manipulation of Diagonal Attention:} In order to verify the  meaning of our diagonal attention maps, 
Figure \ref{fig:map} shows the generated samples by directly manipulating
the diagonal attention maps at specific layers. With 4$\times$4 maps, we can generate the faces with arbitrary direction by changing the activated regions. Also for 8$\times$8 maps, we can control the mouth expression with high values on larger mouth areas. In 16$\times$16 maps, we can control the size of eyes by manipulating activated pixel areas of eyes. These show that our diagonal attention
maps have a clear and intuitive relationship to different
spatial regions. More examples can be found in our Supplementary Materials.

\subsection{Quantitative Comparison Results}


In Table \ref{table:compare}, our model shows better performance in terms of disentanglement metric for almost all of the settings. 
Specifically, when we compare the models trained with both conditions of with and without per-pixel noises, we can see that our models show improved disentanglement metrics compared to  SNI. 
The results clearly indicate that our diagonal attention map can obtain better disentanglement with rich control  of the content than SNI.
Even with the baseline SNI trained with DS loss, the model still could not overcome the limitation of insufficient capacity as indicated by the higher PPL scores.
For further comparison, we also measured the disentanglement of not only the entire $W$ space, but also the style space $W_s$ and the content space $W_c$ each. In all cases, our model achieved improved disentanglement performance with lower PPL scores.
{In addition, our models show comparable FID scores in almost all experimental settings. Although there is a slight degradation in some cases, they are from the expected trade-off between the image quality and the  disentanglement as stated in \cite{sni}.
}
{To support the quantitative results,  qualitative comparison with other methods are provided in Supplementary Materials,
in addition to   extensive ablation studies.}



\begin{table}[!t]
\begin{center}
    \vspace*{-.3cm}
\resizebox{0.4\textwidth}{!}{
\begin{tabular}{c|c|ccc|c}
\hline

\multicolumn{2}{c|}{Per-pixel Noise} &$W$ PPL& $W_s$ PPL& $W_c$ PPL & FID\\
\hline\hline
\multirow{4}{*}{CQ}& StyleGAN& 85.96&-&-&  8.87\\
&SNI& 58.21& 35.35&29.74& 10.79 \\

&SNI+DS&57.63&20.35&31.83&12.10\\
\cline{2-6}
&Ours &\textbf{48.12}&\textbf{18.61}&\textbf{24.19}& 10.90\\
\hline\hline

\multirow{4}{*}{AQ}&StyleGAN& 97.83&-&-& 12.93\\
&SNI & 65.22 & 43.62 & 18.82& 11.32\\

&SNI+DS&69.70&45.79&18.20&15.35\\
\cline{2-6}
&Ours & \textbf{63.44}& \textbf{42.17}& \textbf{17.73}& 11.73  \\
\hline

\hline
\hline
\multicolumn{2}{c|}{w/o Per-pixel Noise} &$W$ PPL& $W_s$ PPL& $W_c$ PPL& FID \\

\hline\hline
\multirow{4}{*}{CQ}&StyleGAN& 112.23&-&-& 9.59\\
&SNI & 70.64&33.11&31.35& 10.93 \\

&SNI+DS&90.81&35.61&45.12&9.89\\
\cline{2-6}
&Ours &\textbf{53.72}&\textbf{21.92}&\textbf{30.15}& 11.40 \\
\hline\hline

\multirow{4}{*}{AQ}&StyleGAN &374.72&-&-&13.92\\
&SNI  & 127.99 & 64.49 & 62.41& 11.91 \\

&SNI+DS&143.22&80.30&50.16&13.52\\
\cline{2-6}
&Ours & \textbf{73.51}&\textbf{38.67}& \textbf{26.83}& 12.67 \\
\hline

\end{tabular}
}
\end{center}
    \vspace*{-.7cm}
\caption{Comparison of FID and PPL scores of models trained using CelebA-HQ and AFHQ datasets at 256$\times$256 resolution. A lower PPL indicates better disentanglement, and a lower FID indicates higher image quality. CQ: CelebA-HQ , AQ: AFHQ, s: Style,c: Content.}
\vspace*{-.5cm}
\label{table:compare}
\end{table}

\subsection{Inverting Disentangled Model}


\begin{figure*}[!ht]
\vspace*{-.3cm}
\centering
  \includegraphics[width=0.95\linewidth]{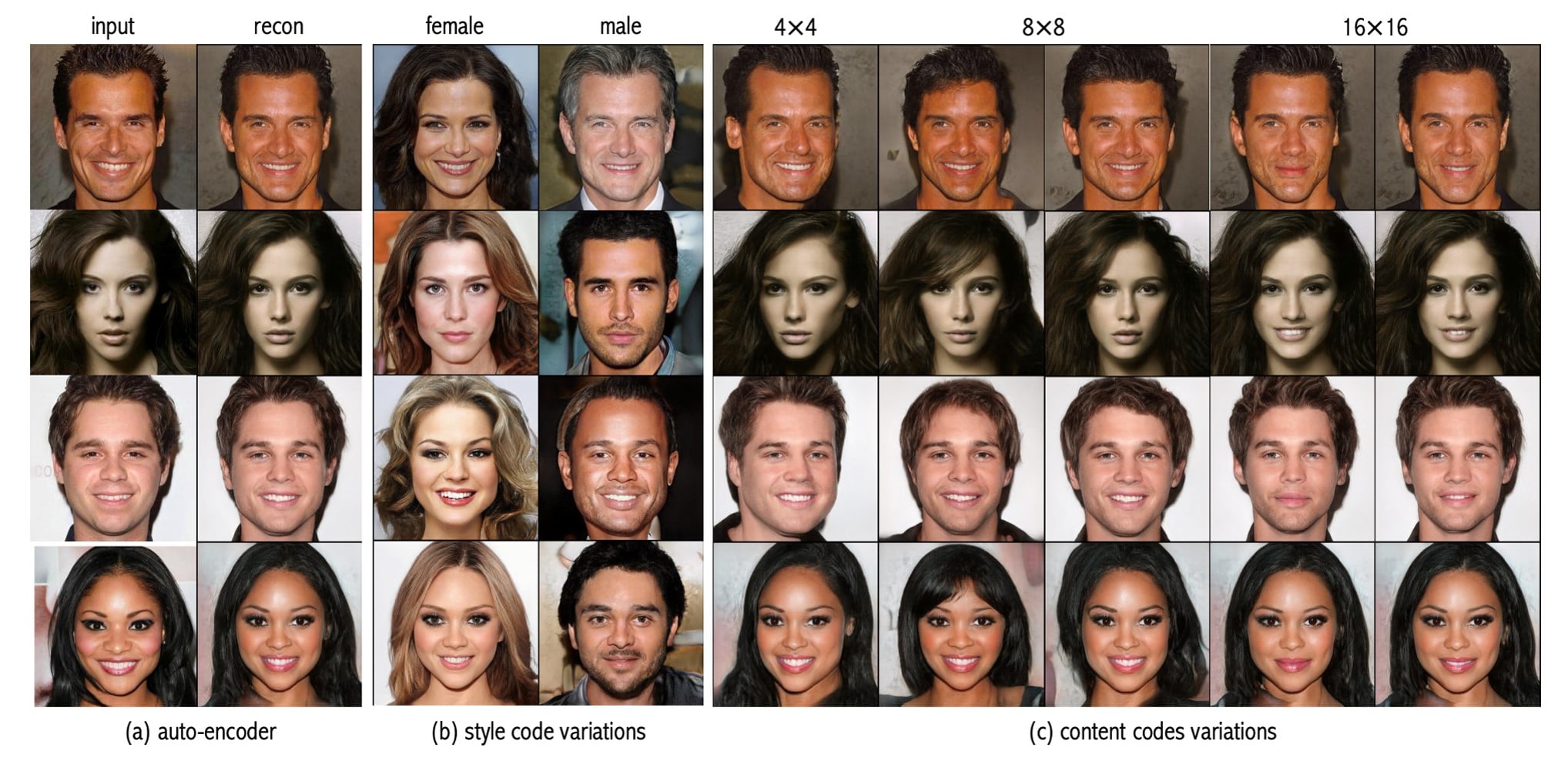}
  \vspace*{-.3cm}
\caption{CelebA-HQ image synthesis results with our inversion model. (a) Auto-encoder reconstruction results from the input.
(b) Images generated using  random style codes by fixing content from the auto-encoder. We can specifically choose the domain (females or males) to translate the style. (c) With fixed style codes from auto-encoder and varying content codes, we can manipulate the content attributes. 
 By changing the content codes of 4$\times$4 layer, we can see the face directional change. By changing the content codes of 8$\times$8 layer, the hairstyle changes. By changing the content codes of 16$\times$16 layer, the mouth expression varies.}
\label{fig:invresult}
\end{figure*}

\begin{figure}[!hbt]
\vspace*{-.5cm}
\centering
  \includegraphics[width=0.8\linewidth]{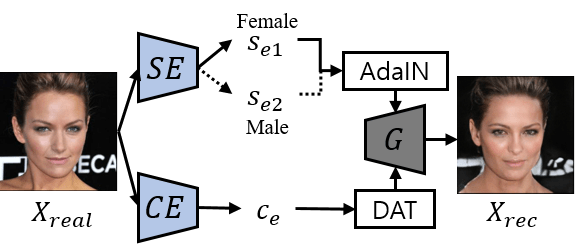}
  \vspace*{-.3cm}
\caption{Network architecture for inverting pre-trained generator $G$. Our style encoder $SE$ network produces multi-domain style codes. Content encoder $CE$ produces domain-invariant content code.}\vspace*{-.3cm}
\label{fig:inv}
\end{figure}

To further highlight the advantage of our method, we additionally implement a GAN inversion framework in which the real images are encoded into latent spaces, from which various output images are generated by simply manipulating content and style codes. 
For realistic image reconstruction, we use the modified version of state-of-the-art inversion method IDinvert \cite{idinvert} to include
both DAT and AdaIN. 

Specifically, we first pretrained our Diagonal GAN with multi-domain styles.
Then, as shown in  Fig.~\ref{fig:inv},  we train the style encoder $SE$ which has a double-head structure so that sampled style codes from each head represent the specific domain style (e.g. male, female). Additionally, the content
encoder $CE$ is trained so that it can generate the content code.
The generated style and contents codes are fed into the pre-trained Diagonal GAN through AdaIN and DAT. Then, we train the network to reconstruct realistic input images. 
%
 For encoder and diagonal GAN network training, we use 28,000 CelebA-HQ images with 256$\times$256 resolution, which are split in two domains of males and females. For testing, we use 2,000 (1000 male, 1000 female) images.
Detailed training process is elaborated in our Supplementary Materials.



Fig.~\ref{fig:invresult} shows the synthesis results from our inversion model. First, auto-encoding reconstruction results confirm that the network can successfully
generate similar outputs as the input images.
Then, Figs.~\ref{fig:invresult}(b) show the results by changing the style codes.
We can change the global styles from the inputs.
In Figs.~\ref{fig:invresult}(c), we show the results by varying the content codes at each resolution layers. 
Thanks to the DAT layers,  compared to the existing  image translation models,
our model has much more flexibility by allowing hierarchically control of both content and styles in the generated images.


\begin{table}[!hbt]
\begin{center}
 \begin{adjustbox}{width=0.35\textwidth}
\vspace{-0.3cm}
\begin{tabular}{|c|cc|cc|}
\hline

{Methods} &\multicolumn{2}{c|}{Latent}& \multicolumn{2}{c|}{Reference}\\
\cline{2-5}
 &FID & LPIPS & FID & LPIPS\\
\hline\hline
StarGANv2& 13.05& \textbf{0.453}&22.35&0.405\\
Ours& \textbf{11.12}&0.452& \textbf{18.11}& \textbf{0.407}\\
\hline
\end{tabular}
 \end{adjustbox}
\end{center}
    \vspace*{-.5cm}
\caption{Quantitative comparison results of style synthesis using GAN inversion with CelebA-HQ. Lower FID means better image quality, and higher LPIPS means more diversity.}
\label{table:inv}
\end{table}

For further evaluation, in Table \ref{table:inv}, we compared the performance with the state-of-the-art image translation model StarGANv2 \cite{starganv2}. Since the existing  StarGANv2 can only change the style similar to Figure \ref{fig:invresult}(b), we measured the quantitative performance of style synthesis for fair comparison. Surprisingly, we achieved better image quality with comparable diversity even in style synthesis for both of latent-based sampling and reference-based transfer. The results show that our method has remarkable advantages, as it has a better image generation quality and more flexible content control than the existing state-of-the-art model. Detailed experiment settings and qualitative comparisons are provided in Supplementary Materials.


\section{Conclusions}
In this paper, we proposed  a novel diagonal spatial attention (DAT) module as a complement to the AdaIN in order to
 disentangle the style and content information. 
The symmetric structure of DAT and AdaIN enabled the independent control of  the style and content of features in a hierarchical manner.  
Our extensive experiments  showed that
 the style and content attribute of images can be independently manipulated in a hierarchical manner, confirming the style and content
 disentanglement in high quality image generation.
  Moreover, the proposed method has also been successfully integrated into GAN inversion to achieve high quality image translation with better disentanglement of content and style.

%


\section*{\Large{\textbf{Supplementary Material}}}

\setcounter{section}{0}
\renewcommand*{\theHsection}{chX.\the\value{section}}

\section{Diagonal GAN Experiments}
\subsection{Qualitative Experiment Setting}
For qualitative evaluation in the main text,  our models were trained using the images from 1024$\times$1024 CelebA-HQ \cite{progan} and 512$\times$512 AFHQ \cite{starganv2}. 

Specifically, using full-resolution CelebA-HQ images, we trained our model by accessing 20 million training samples. 
For efficient training with limited GPU capacity, we started with batch size of 512 in the first 8$\times$8 resolution, and reduced the batch size by half each time when we proceeded to a larger image size. With the aforementioned training strategy, the overall training took about one month using a single Tesla V100 GPU. For learning rate, we used 0.001 until accessing 12 million samples, then decreased the learning rate to 0.0001.  
To test the effect of our diagonal attention (DAT) module in the qualitative evaluation, we  removed per-pixel noises at each layer. 
We increased the $\lambda$ value of DS loss from 0.3 to 0.5 after accessing 12 million samples for the training. 
We used the left and right flips for data augmentation in all training procedures.
For the case of full resolution AFHQ images, we used the same training settings as we did before for the CelebA-HQ dataset case, except that we used the fixed value of $\lambda$ as 0.3 throughout the training.

For further qualitative evaluations, we carried out experiments with additional image data sets: Oxford Flowers 102 \cite{flower}, Caltech-UCSD Birds (CUB2011) \cite{bird}, and Stanford Cars \cite{car}. For the flower data set, we first extracted the flower regions with center cropping. Then we resized the cropped images to 512$\times$512. For the bird dataset, we extracted bird image areas using  bounding box information. Then we changed the size of the extracted images to 256$\times$256. For the car dataset, we also extracted the car image areas using bounding box information, then resized the cropped images to 384$\times$512. We also used the left-right flips for data augmentation.

When training the models with flower and car data sets, we continued the training models with up to 12 million samples.
For the bird dataset,  training was continued until we  accessed  10 million samples. 
The  {training settings} used for the AFHQ model training was also used for the flower, car, and bird datasets.

To improve the perceptual quality, the images  are generated by applying truncation trick similar to \cite{biggan}. 
More specifically, we found that best perceptual image quality was obtained by
 truncating mapped style code in $\W_s$ up to 0.7, whereas no truncation was used for $\W_c$. 

\subsubsection{Additional Qualitative Experiment Results}

\noindent\textbf{AFHQ results:} Fig.~\ref{fig:af_map} shows the results of direct attention map manipulation of  our full-resolution AFHQ model. Similar to our CelebA-HQ results, we could obtain the faces with the desired direction by manipulating the 4$\times$4 map, and 
control the mouth opening by changing the values around mouth in 8$\times$8 map.
We also show the results of the interpolation of the content codes of our AFHQ model in Figs.~\ref{fig:f0} and \ref{fig:f02}. When the content codes at all levels are changed, the global spatial attributes are changed, and when the 8$\times$8 maps are changed, the lower parts of the areas change. Quantitatively, our model trained with AFHQ in full resolution achieved 10.79 in FID.


\noindent\textbf{CelebA-HQ results:} We also show more results of the content code interpolation with the model trained with full-resolution CelebA-HQ in Figure \ref{fig:c1} and \ref{fig:c2}. When the content codes for all layers are changed, the global spatial attributes are changed, and when the first 4$\times$4 codes are changed, face direction changes.  When the 8$\times$8 codes are changed, lower parts of faces change. 

\noindent\textbf{Flower results:} To test the versatility of our proposed model, we trained our  model using additional datasets. Fig.~\ref{fig:f1} illustrates the generated images from the model trained on flower dataset. When we change the content code with fixed style codes, we can change the spatial information such as flower shape, number, and location of flowers.
On the other hand,  if we vary style codes with fixed content codes, we can observe the changes of the global style attributes including species, flower color and background. 
Fig.~\ref{fig:f2} also shows the results by changing the content smoothly with interpolating the content codes. Our flower model scored 46.43 in FID.

\noindent\textbf{Birds results:} Fig.~\ref{fig:c4}  shows the generated images from the model trained on birds dataset. 
Fig.~\ref{fig:c4} (b) shows samples with varying style codes and the fixed content code. We can observe the changes of  the global style attributes including species, feather colors and patterns. Fig.~\ref{fig:c4} (c) illustrate samples generated with varying content codes and the fixed style. We can observe that the content attributes including location, rotation and global shapes change with different content codes.
Fig.~\ref{fig:c5} also shows the content interpolation results, which shows that birds smoothly change the head orientation.
%
 Our model on birds dataset scored 14.27 in FID.

\noindent\textbf{Cars results:} Figure \ref{fig:c6}(a) shows the sample images from random content and style code.
Then, Figure \ref{fig:c6}(b) shows the samples with varying style codes and the fixed content code. We can observe  the changes of the global style attributes including car type, colors and background. Samples generated with varying content codes and the fixed style are provided
in Figure \ref{fig:c6}(c). We can observe that the content attributes including rotation and global shapes change with different content codes.
In Fig.~\ref{fig:c7}, we can see the effect of content interpolation in terms of rotation angle.
 Our model on car dataset scored 8.96 in FID.

\subsection{Quantitative Experiments}
For quantitative evaluation, we compared our method with the baseline SNI model, SNI with DS loss, and the original StyleGAN. In order to carry out extensive comparative studies with various models, we trained the models with the reduced resolution of $ 256 \times  256$ using 500,000 iterations (total of $\sim$4.7 million samples). We also used batch-size scheduling for efficient training. It took four days for training each model with a single NVIDIA RTX2080Ti GPU. For a fair comparison, we used the same non-saturating loss with {\it{R$_1$}} regularization in all the experiments. 
The same settings are also used in all models for our ablation studies and additional disentanglement studies.

For training the baseline SNI  model and SNI with DS loss, we implemented the models on PyTorch based on official source code\footnote{https://github.com/yalharbi/StructuredNoiseInjection}. To follow the best settings in the original paper \cite{sni}, we used the input tensor of 8$\times$8 resolution for training the models. For the DS loss, $\lambda$ is set to 0.3, which shows the best results.

In order to quantitatively evaluate the image quality of the generated samples, we calculated the FID values  \cite{fid}. For CelebA-HQ with 30,000 training images, we computed the FID values with 50,000 generated samples. For the AFHQ and other data sets with relatively fewer training images, we calculated the FID values with 20,000 generated samples.

To calculate the total PPL of the $W$ space, we follow the same calculation proposed in StyleGAN\cite{stylegan}. If we sample the two style codes $s_1,s_2\in W_s$ and two contents codes $c_1,c_2\in W_c$, the PPL score is calculated as
\begin{align*}
   & \text{PPL}_W \\
  = & \frac{1}{\epsilon^2}\mathbb{E}\left[d(G(ts_1+(1-t)s_2,t c_1+(1-t)c_2)),\right.\\
  & \left. G((t+\epsilon)s_1+(1-t-\epsilon)s_2,(t+\epsilon) c_1+(1-t-\epsilon)c_2))\right]
\end{align*}
where $t$ is a uniformly sampled between $[0,1]$, and  $G(s,c)$ is the generator output with respect to the style code $s$ and content code $c$, respectively, and $d(X,Y)$ denotes the perceptual distance between two images $X$ and $Y$. We use $\epsilon = 10^{-4}$ for all the calculations
and report the average values that are computed using 10,000 generated samples.

For calculation of PPL for $W_s$, we use fixed content code $c_{fix}\in W_c$ and paired style codes $s_1,s_2\in W_s$:
\begin{align*}
 & \text{PPL}_{W_s} \\
  = & \frac{1}{\epsilon^2}\mathbb{E}\left[d(G(ts_1+(1-t)s_2, c_{fix})),\right.\\
  & \left. G((t+\epsilon)s_1+(1-t-\epsilon)s_2,c_{fix})\right]
  \end{align*}
To calculate $\text{PPL}_{W_c}$, we use fixed style code $s_{fix}\in W_s$ and sampled content codes $c_1,c_2\in W_c$:
\begin{align*}
 & \text{PPL}_{W_c} \\
  = & \frac{1}{\epsilon^2}\mathbb{E}\left[d(G(s_{fix},tc_1+(1-t)c_2),\right.\\
  & \left. G(s_{fix}, (t+\epsilon)c_1+(1-t-\epsilon)c_2)\right]
\end{align*}
For the computation of PPL for $W_s$ (resp. $W_c$),
for each fixed code,  the content codes (resp. style codes)
are sampled fifty times,  and  the average value was calculated by repeating this 200 times.
 Therefore, the final PPL value is calculated using 10,000 samples.

\subsection{Ablation studies}

\begin{table}[!t]
\begin{center}
\begin{adjustbox}{width=0.35\textwidth}
\begin{tabular}{@{\extracolsep{\fill}}c|cc|cc}
\hline
\multirow{2}{*}{{Varying $\lambda$}}& \multicolumn{2}{c|}{CelebA-HQ} & \multicolumn{2}{c}{AFHQ}\\
\cline{2-5}
&FID&$W$ PPL&FID&$W$ PPL\\
\hline
\hline
0.2& 11.65&49.83 &13.27 & 70.78
\\
0.3 & \textbf{10.90}& \textbf{48.12}
& \textbf{11.73}& \textbf{63.44}
\\
0.4& 11.69&62.44
& 13.32& 80.41
\\
0.5& 11.51& 68.75
& 13.87& 73.12
\\


\hline
\end{tabular}
\end{adjustbox}
\end{center}

\caption{Quantitative results of ablation study. We investigate the effect of $\lambda$ value in the diversity-sensitive loss. }
\label{table:lambda}
\end{table}

\begin{table}[!t]
\begin{center}
 \begin{adjustbox}{width=0.4\textwidth}
\begin{tabular}{@{\extracolsep{\fill}}c|cc|cc}
\hline
\multirow{2}{*}{DAT network}& \multicolumn{2}{c|}{CelebA-HQ} & \multicolumn{2}{c}{AFHQ}\\
\cline{2-5}
&FID&$W$ PPL&FID&$W$ PPL\\
\hline
\hline
2$\times$MLP-256&14.51& 53.99&16.85&70.02 \\
CNN-256& 11.18&83.37& 14.28&91.88\\
single MLP-32&11.74& 60.87&13.26&118.21 \\
single MLP-64& \textbf{10.66}&61.98& 12.50&87.89\\
single MLP-256 &10.90&\textbf{48.12}&\textbf{11.73}&\textbf{63.44}\\
\hline
\end{tabular}
 \end{adjustbox}
\end{center}

\caption{Quantitative results of ablation study. We compare different attention mapping networks and the maximum resolution for the DAT layers.}
\label{table:map}
\end{table}


{In our ablation study, we compared the quantitative performance of models trained with different settings. In all of the experiments, we used models trained with per-pixel noises.}


In order to validate the choice of the value of $\lambda$ for the diversity-sensitive loss, we first show the results of various models that were trained with different $\lambda$. In Table~\ref{table:lambda}, the models trained with $\lambda = 0.3$ show the best performance 
 in the disentanglement capability, exhibiting  the lowest PPL scores; furthermore, they showed the best image quality with the lowest FID. The models trained with lower or higher $\lambda$ values show degraded disentanglement performance. The results show that we can achieve the most balanced content-style control with both codes when we set $\lambda=0.3$. 

We also investigated the effect of different network architectures for the attention mapping, and show the results in Table~\ref{table:map}. 
To ensure stability in training, we use attention mapping with a single layer MLP followed by a sigmoid applied to layers with a resolution of up to $256 \times 256$.
To verify our choice of mapping network architecture, we implemented two additional networks for ablation study: {{2$\times$MLP-256}} and {{CNN-256}}. Here, {{2$\times$MLP-256}} represents a model which has attention mapping of 2-layer MLP instead of a single MLP. 
The model {{CNN-256}} uses CNN layer-wise upsampling network to generate the diagonal attention. In all the experiments, we fixed $\lambda=0.3$ and used mapping network up to 256$\times$256 layers. 

Table~\ref{table:map} shows that when using 2-layer MLP, we can obtain a well-disentangled model with relatively low PPL scores, but it still cannot achieve the best performance. In case of using CNN as an attention network, the disentanglement scores are severely degraded, which may be due to the imbalance between the simple AdaIN network and CNN-based  attention mapping networks. 

Then, we carried out comparative study
by changing the maximum resolution of the diagonal attention (DAT) layer.
In Table \ref{table:map}, the use of DAT up to 32$\times$32  (single MLP-32) and 64$\times$64 (single MLP-64) have relatively high PPL values, suggesting that DAT layers at the lower levels only result in a limited expressiveness for the various content information.  The disentanglement quality is particularly impaired in the models trained with AFHQ data set. We suspect that limited capacity in content control makes it more difficult to cover the variations of images in AFHQ that are more diverse than those in CelebA-HQ. Therefore, we use DAT layers up to 256$\times$256 resolution (single MLP-256), which is our default model.

 In evaluating image quality in terms of FID scores, our default model showed better performance than most of baseline settings except for single MLP-64. However, in the case of single MLP-64, the disentanglement performance is relatively poor. Therefore, we can obtain best result when using single MLP-256.

\subsection{Disentanglement Experiments}
\begin{table}[!t]
\begin{center}




\resizebox{0.47\textwidth}{!}{
\begin{tabular}{c|c|ccc|ccc}
\cline{1-8}
	\multirow{2}{*}{Dataset}&\multirow{2}{*}{Methods}& \multicolumn{3}{c|}{Per-pixel Noise} & \multicolumn{3}{c}{w/o Per-pixel Noise}\\ \cline{3-8}
	&& content& style & all& content& style & all\\
	\cline{3-8}
\hline \hline

\multirow{3}{*}{CelebA-HQ} &SNI \cite{sni}&0.338&0.499&0.532 &0.348 & 0.496 & 0.534\\
&SNI+DS&0.392&0.472&0.531 &0.378 & 0.489 & 0.535\\
&Ours&0.411&0.469&0.532&0.413&0.472&0.533\\
\hline
\multirow{3}{*}{AFHQ} &SNI \cite{sni} &0.344&0.588&0.605&0.411&0.583&0.610\\
&SNI+DS&0.399&0.585&0.607 &0.407 & 0.589 & 0.608\\
&Ours&0.412&0.582&0.607&0.443&0.578&0.608\\
\hline
\end{tabular}
}
\end{center}
\caption{Comparison of LPIPS scores of models trained using CelebA-HQ and AFHQ datasets at 256$\times$256 resolution. Our model has more balanced content-style control than that of the baseline models.}
\label{table:lpips}
\end{table}



To further quantify the disentanglement performance,
we additionally  measure  the content and style diversity in the image generation. As a measurement of image diversity, we use Learned Perceptual Image Patch Similarity (LPIPS) \cite{LPIPS}. Since our model and the baseline SNI have two independent content and style codes, we can compare their diversity of style and content. To measure the diversity score of both codes, we compute the average value of LPIPS of 40,000 images sampled with arbitrary content and style codes. On the other hand, to measure the style and content diversity separately, we calculate the LPIPS of 40 images sampled by varying one code with another code fixed, which is repeated for 10,000 times to calculate the averaged LPIPS. 
This makes the the total number of images for LPIPS calculation equal to  40,000 for both cases.

Table \ref{table:lpips} is the result of LPIPS scores. When we compare the LPIPS by varying both content and style codes, both  SNI and our model show similar diversity in the generated images.
However, when looking at the diversity of style and content separately, the baseline SNI shows that the diversity of content is much lower than that of the style. On the other hand, our model has more balanced diversity in style and content. With SNI trained with DS loss on content code, the model shows slightly better diversity in content code than that of the baseline SNI. However, the model still shows lower content diversity than our model as it is not able to overcome the capacity limit of content control with input tensor. 

To support the above quantitative comparison in terms of  LPIPS scores, we  also qualitatively compared the effect of various content controlling methods: per-pixel noises of StyleGAN, input tensor of SNI and our DAT mapping. 
Since the code spaces of the baseline models are slightly different, we compare the images generated from the mean style code by varying the content codes.
In Figure \ref{fig:compare_sup}, we can see that  StyleGAN with different per-pixel noises only results in  the minor spatial variations such as curls of hair and fur. As for the baseline SNI, it can only change the simple geometrical information such as rotation. For SNI trained with DS loss, we can see that the generated samples have more diversity than basic SNI, but still the variation is limited to geometry similar to basic SNI. In contrast, our model allows more diverse changes on spatial information including geometry, hairstyle (fur pattern), facial expression, etc., by controlling specific DAT layers.



\section{Diagonal GAN Inversion}
\subsection{Methods}
\begin{figure}[!t]
\centering
  \includegraphics[width=\linewidth]{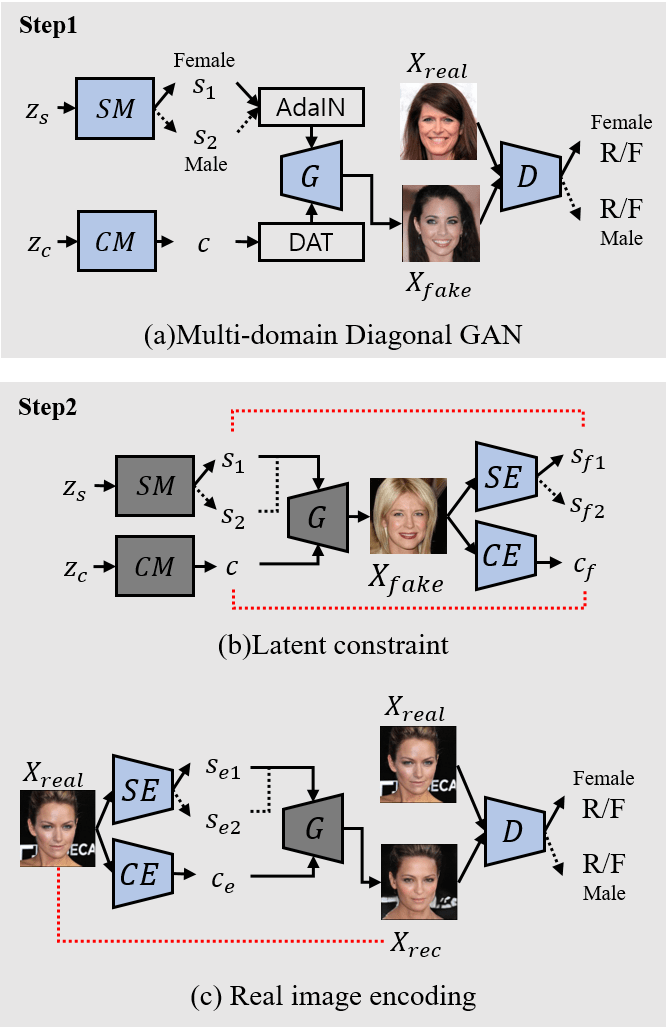}
\caption{Detailed training procedure for GAN inversion task. Step 1: Training our Diagonal GAN with multi-domain style codes. Step 2: With pre-trained Diagonal GAN in Step 1, we introduce style encoder $SE$ and content encoder $CE$ to find style and content codes for real input images. The networks with gray color are fixed, and with blue color are trainable. Red dotted lines indicate supervision losses.}
\label{fig:invfull}
\end{figure}

As discussed in the main text, our diagonal GAN can be easily incorporated with GAN inversion.
Specifically, our inversion model consists of two steps as shown in Figure \ref{fig:invfull}.  The details of each step
are as follows.

\noindent\textbf{Step 1:} The first step is to train our proposed Diagonal GAN. For domain-aware (i.e females, males) image generation,
we train a multi-domain Diagonal GAN in which the style mapping network $SM$ can sample multiple style codes with a multi-head structure. 
Specifically, we use two types of  style codes that represent males and females domains (see  Fig.~\ref{fig:invfull}(a)). On the other hand,
the content mapping generates a unified content code $c = CM(z_c)$ that can be used for both style domains. We used mapped style codes $s\in W_s$ which have the dimension of 512, and mapped content codes $c\in W_I$ with the dimension of 512. Our discriminator $D$ also has a multi-head structure to simultaneously enable realistic generation and domain classification. 

\noindent\textbf{Step 2:} After pre-training our generator network (Fig.~\ref{fig:invfull}(a)), we invert the real images into latent spaces with our inversion network in Step 2 (Figs.~\ref{fig:invfull}(b)(c)). In this step, we use the modified version of state-of-the-art GAN inversion method IDinvert \cite{idinvert}. To encode the real images into style and content code spaces, we introduced the style encoder $SE$ and the content encoder $CE$ networks. Similar to the style mapping in Step 1, our style encoder has a multi-head structure after the last convolution layer.  

The main idea of IDInvert is that when encoding a real image into a latent space, realistic reconstruction is possible only when the encoded latent code is constrained within the learned latent space. To achieve this goal, as shown in Fig.~\ref{fig:invfull}(b), we first sampled the random style code $s$ with random domain label $y$, and the random content code $c$ using the pre-trained mapping networks $SM$ and $CM$, and generated a fake image $X_{fake}$. Then by putting the generated $X_{fake}$ into the encoders, we can obtain the encoded style and content codes $s_f$ and $c_f$, respectively. Then, our loss for latent codes is given by
\begin{equation}
    L_{latent} = ||s-s_f||_2 +||c-c_f||_2
\end{equation}
which  reduces the mean-squared error (MSE) between the encoded codes and the learned codes so that our encoder networks can generate the codes within the learned latent spaces. 

Additionally,  in Fig.~\ref{fig:invfull}(c), we put the real image $X_{real}$ with the corresponding domain label $\hat{y}$ into the style and content encoders to get the content code $c_e$ and style code $s_e$, respectively. Then, the codes $c_e$ and $s_e$ are used in DAT and AdaIN layers, respectively,  of the pre-trained generator to obtain the reconstructed image $X_{rec}$. The goal of this step is 
to make $X_{rec}$ as close as possible to $X_{real}$. For realistic reconstruction, we reduce the distance between $X_{real}$ and $X_{rec}$ by using a MSE loss, a LPIPS \cite{LPIPS} loss that reduces the perceptual distance, and an adversarial loss using a new discriminator $D$ which also has a multi-head structure. For adversarial loss, we used the same loss function as StyleGAN \cite{stylegan}, which is composed of the non-saturating Softplus, $f(t) = \text{softplus}(t) = \text{log}(1+\text{exp}(t))$, with $R_1$ regularization. 

Accordingly, our total loss function for the content and style encoder is given by
\begin{align*}
  L_{E} =&  ||X_{real}-X_{rec}||_2 
    +LPIPS(X_{real},X_{rec}) \\
 &   +\lambda_{lat}L_{latent} + \lambda_{adv} f(-D_{\hat{y}}(X_{rec}))
\end{align*}
where $\lambda_{lat}$ and $\lambda_{adv}$ are weight parameters.
On the other hand, the loss for the discriminator is 
\begin{align*}
\begin{split}
 L_{D} =&  f(D_{\hat{y}}(X_{rec})) + f(-D_{\hat{y}}(X_{real}))  \\
  &  +\frac{\gamma}{2} \mathbb{E}[||\nabla D_{\hat{y}}(X_{real})||^2_2]
\end{split}
\end{align*}
where $D_{\hat{y}}(\cdot)$ denotes the output of the discriminator $D$ corresponding to the domain ${\hat{y}}$,
 $\gamma$ is a weight parameter for gradient $R_1$ regularization (the last term in $L_D$).

\noindent\textbf{Inference time Latent-Regularized Optimization:}
Additionally,
we try to find better latent codes at the test time by additionally optimizing the latent code for better reconstruction. In this process, we use the latent optimization method proposed by IDinvert \cite{idinvert}. Specifically,
as an initialization for  the style and content codes $s$ and $c$, respectively,
 we use codes $s_e$ and $c_e$  from pre-trained encoders in Step 2. In addition to reducing the distance between the input and reconstruction, we include latent regularization loss to make the latent vectors $s$,$c$ lie within the learned space of the encoders and the generator. 
The resulting loss function for optimization is:
\begin{equation}\label{eq:lat_opt}
\begin{split}
L_{code}(s,c) &= ||X_{real}-G(s,c)||_2  +  LPIPS(X_{real},G(s,c)) \\
&+ \lambda_{reg}||s-SE_{\hat{y}}(G(s,c))||_2 \\
&+ \lambda_{reg}||c-CE(G(s,c))||_2
\end{split}
\end{equation}
where $G(s,c)$ denotes the generator output with style and content codes $s$ and $c$, respectively,
 $SE_{\hat{y}}$ refers to the style encoder on the domain $\hat y$,
and $CE$ is the content encoder.

\subsection{Method Details}
In GAN inversion experiments, we used 256$\times$256 resolution CelebA-HQ dataset. Total of 30,000 images, 28,000 are used as a training set, and 2,000 are used as a test set. The test set consists of 1000 male and 1000 female face images.

 When training our Diagonal GAN network in Step 1, we trained the model until we access a total of 10 million training samples, which took about a week with a single RTX-2080Ti GPU.  Except for the maximum resolution, other training settings are the same as our full-resolution CelebA-HQ experiments.

Our style encoder model is a CNN with multi-head fully-connected layers, which has the same structure as the discriminator. The content encoder  has the same architecture, except that it has a single-head structure. In  Step 2 training, we trained the model with the batch size of 2 for 200,000 iterations. We used Adam optimizer, initially using a learning rate of 0.001, and then decreased the learning rate to 0.0001 after 100,000 iterations. For weight parameters, we set $\lambda_{lat}=1$, $\lambda_{adv}=0.1$. This took 2 days with a single NVIDIA RTX2080Ti GPU.

{For our inference time latent-regularized optimization using \eqref{eq:lat_opt},  both style and content latent codes were optimized using 100 iterations per a single input image. We used Adam optimizer with learning rate of 0.01, and set the loss weights as $\lambda_{reg}=2$. Optimization process took 4 seconds per each  input image.}

\subsection{Experiment Details}

In order to show the superior disentanglement performance of our model, we compared our method with state-of-the-art diverse image translation model, StarGANv2 \cite{starganv2}. 
Note that  our inverted model can control both content and style spaces using DAT and AdaIN layers, whereas StarGANv2 can only convert styles of input images due to the exclusive use of AdaIN layers. 
For a fair comparison, we used the pre-trained StarGANv2 that can be downloaded from the official GitHub repository \footnote{https://github.com/clovaai/stargan-v2}. 


For quantitative evaluation, we measured the quality in terms of FID and diversity through LPIPS. 
Since StarGANv2 can only convert the style of the images, we only consider the style conversion by two methods for this quantitative comparison.
We consider both image translation scenario: 1) latent-based image translation, which converts the style of input image to a random style by sampling the style codes, and 2) reference-based image translation, in which we convert the style of inputs to that of the reference images. At this time, we measured the performance by converting a single image of one domain into 10 different target domain images. In our GAN inversion,
 the experiment was conducted by varying the style codes while using  the same content code of the input image. 
 As mentioned before, this is to compare the image quality during style translation, as StarGANv2 is only for the style translation.
 Since the test set contains 1,000 images  for each domain (female, male) and 10 target styles are used for each {image}, 10,000 synthesized images can be obtained.
 Furthermore,  we consider the domain conversion scenario (i.e. females $\leftrightarrows$ males), which doubles the number of synthesized images.  Therefore, for each latent and reference based experiment, we measured metrics on 20,000 generated images.  For more details, please refer to the original StarGANv2 paper \cite{starganv2}, as we use the same evaluation process.

{In all the experiments, we used style and content codes obtained using inference time latent-regularized optimization process, except for the qualitative experiment of reference-based style synthesis, where style codes without latent-regularized optimization still provide better perceptual quality.}

\subsection{Inversion Experimental Results}



\noindent\textbf{Auto-encoder Reconstruction Results: }
Our inversion model showed satisfactory reconstruction performance by extending the state-of-the-art inversion model.
To evaluate the reconstruction performance, we measured the distance between input and the reconstructed image with MSE and LPIPS. When we reconstruct the images without inference time latent-regularized optimization, we could obtain MSE of 0.095 and LPIPS of 0.246. Furthermore, with additional latent-regularized optimization process, the model showed improved performance with MSE of 0.042 and LPIPS of 0.155. 
The results confirmed that our model shows good reconstruction performance, and more accurate reconstruction is possible when latent-regularized optimization is additionally used.

\noindent\textbf{Qualitative comparison:} In the main script, we have already compared the performance of our model and baseline StarGANv2 to show that our model outperforms the generation quality. Here, we provide
more extensive qualitative comparison results to highlight  the advantages of our model.

Figure \ref{fig:compare_reference} shows the generated samples synthesized from input image to follow the styles of reference images. Although StarGANv2 shows good performance in style synthesis from typical images (see Fig.~\ref{fig:compare_reference}(a)), it is still a conventional image translation model that uses spatial information of the image as it is. Therefore, as shown in Fig.~\ref{fig:compare_reference}(b), 
when the content information of the input is complicated or rare, we can observe that the generation performance is often  severely degraded. In contrast,  our model finds the content code that can best express the input content in the pre-trained space, so that it can generate realistic images even with complex or rare input contents (see Fig.~\ref{fig:compare_reference}(b)).
Figs. \ref{fig:compare_latent}(a)(b) show the result of converting the input image to follow random styles. Again, our model could generate more realistic images even if the input content is complex (see (see Fig.~\ref{fig:compare_latent}(b)).

To clearly show the strength of our model, in Figure \ref{fig:compare_intp}, we show the results from our image translation
results with content manipulation through the content code interpolation. As explained before, StarGANv2 can change the contents only by using different input images similar to Figs.~\ref{fig:compare_intp}(a)(c). In contrast, as our model can control the content code space,
 Figure \ref{fig:compare_intp}(b) shows that the content information of  the translated images can change smoothly
with interpolating the content codes.
Furthermore,  Figure \ref{fig:compare_content} shows the image translation experiment by changing the hierarchical content code in addition to style synthesis. Unlike StarGANv2, which can only change the style of the input, we can further change the the specific content attributes such as: face geometry by changing the 4$\times$4 content codes, the hair shape by changing the 8$\times$8 codes, and the mouth expression by changing the 16$\times$16 codes.

The results clearly show that our model is capable of flexible content control  in addition to the style control,
which is not possible with the existing image translation models such as StarGANv2.

{\small
\bibliographystyle{ieee_fullname}
\bibliography{egbib}
}

\newpage

\begin{figure*}[!h]
\centering
\includegraphics[width=0.6\linewidth]{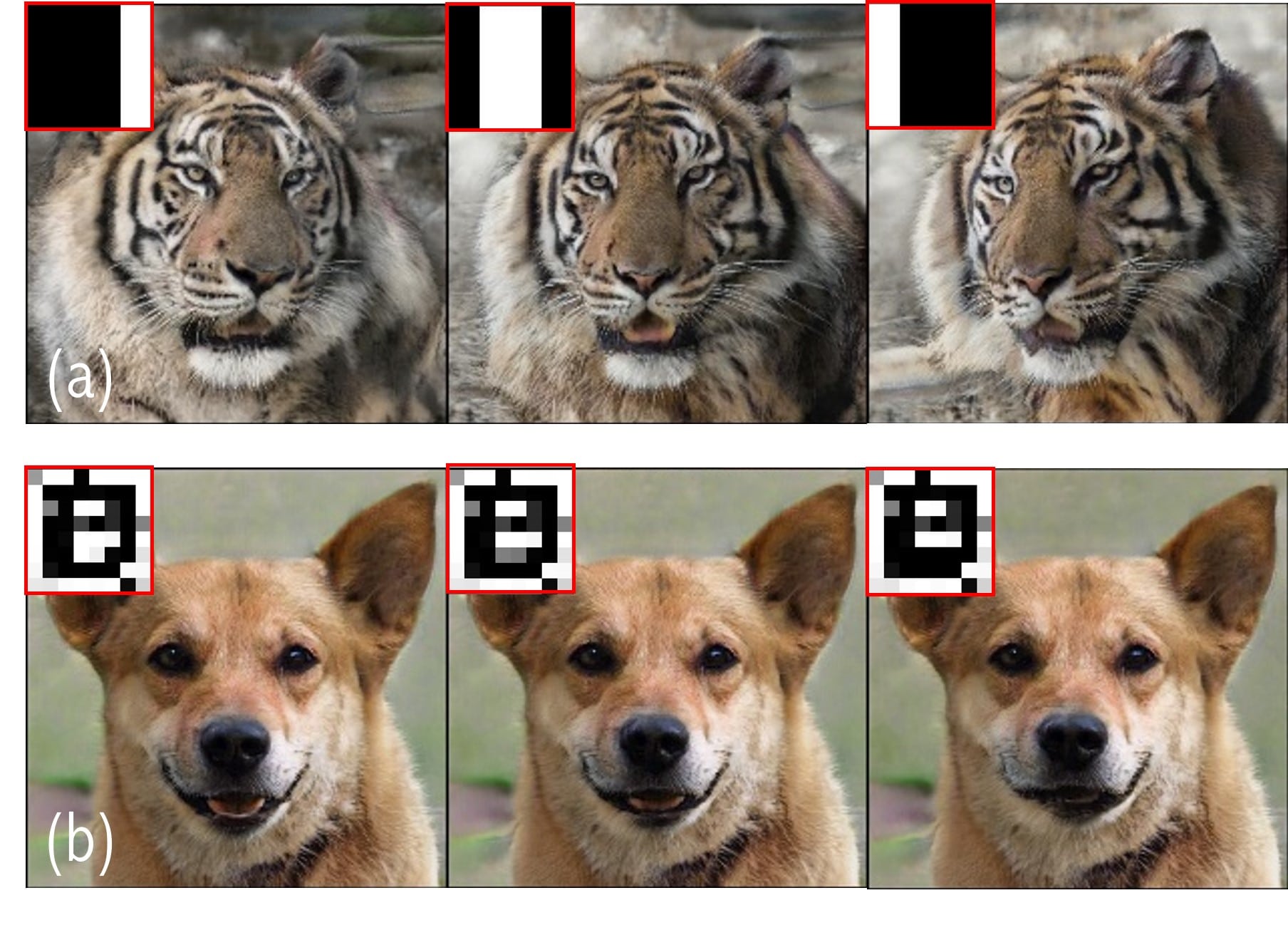}
\vspace*{-0.5cm}
\caption{Direct  attention  map  manipulation. By controlling  the  specific  areas  of  attention,  we  can  selectively change the animal facial attributes.  Results from changing (a) the first 4$\times$4 attention map, (b) the 2nd 8$\times$8 attention map. We can manipulate the face direction with changing first 4$\times$4 map, and change mouth expression with 2nd 8$\times$8 map.}
\label{fig:af_map}
\end{figure*}

\begin{figure*}[!h]
\centering
\includegraphics[width=0.96\linewidth]{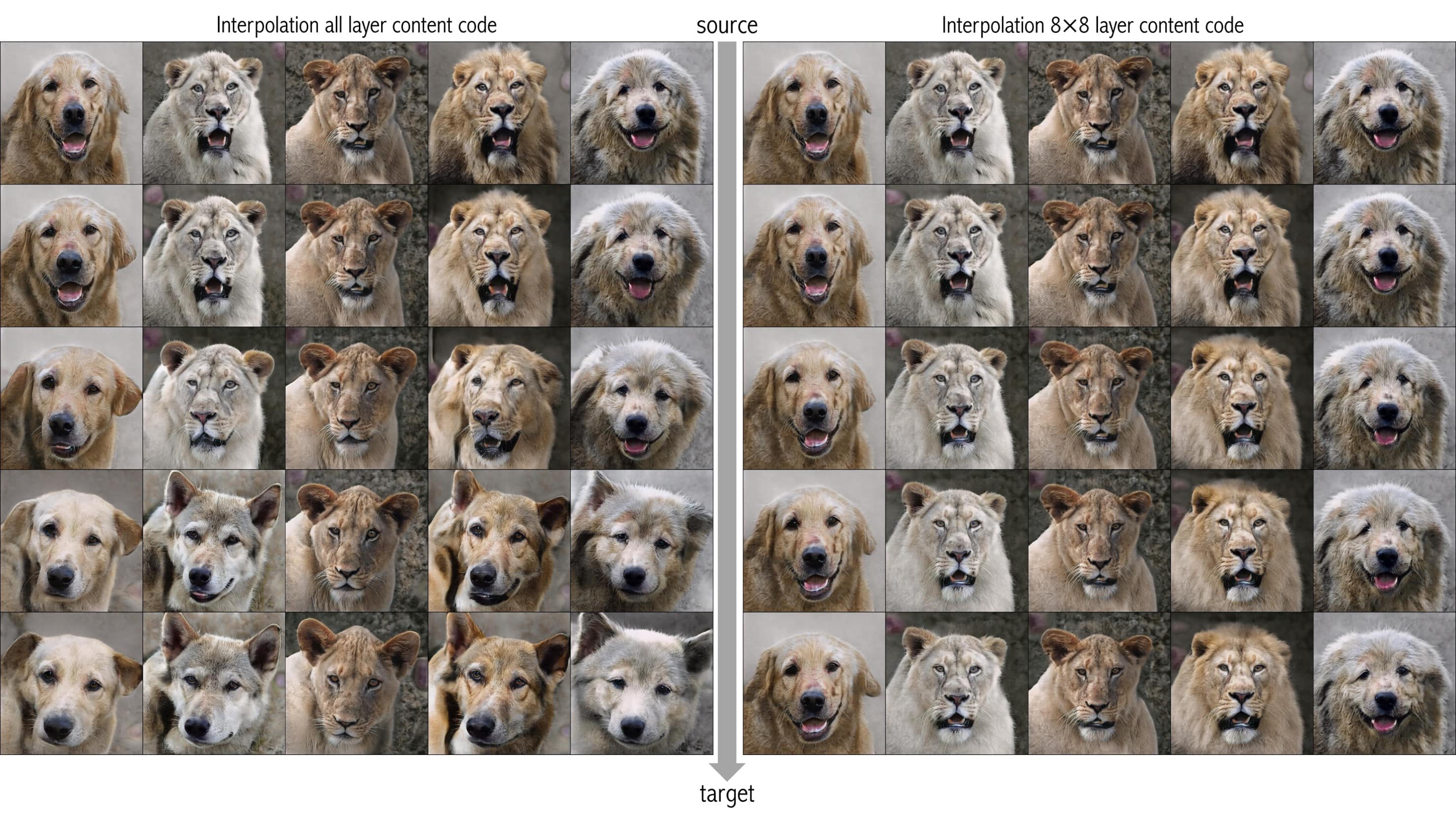}
\vspace*{-0.5cm}
\caption{Examples from interpolated content codes. The images of each row are sampled from the same content code. (Left) Samples  by varying content codes across all layers. We can observe that the global attributes such as rotation, and shape change gradually. (Right) Samples by varying 8$\times$8 attention maps. We can observe that this layer mainly contributes on the lower part of faces (i.e mouth).}
\label{fig:f0}
\end{figure*}
\vfill
\begin{figure*}[!t]
\centering
\includegraphics[width=0.96\linewidth]{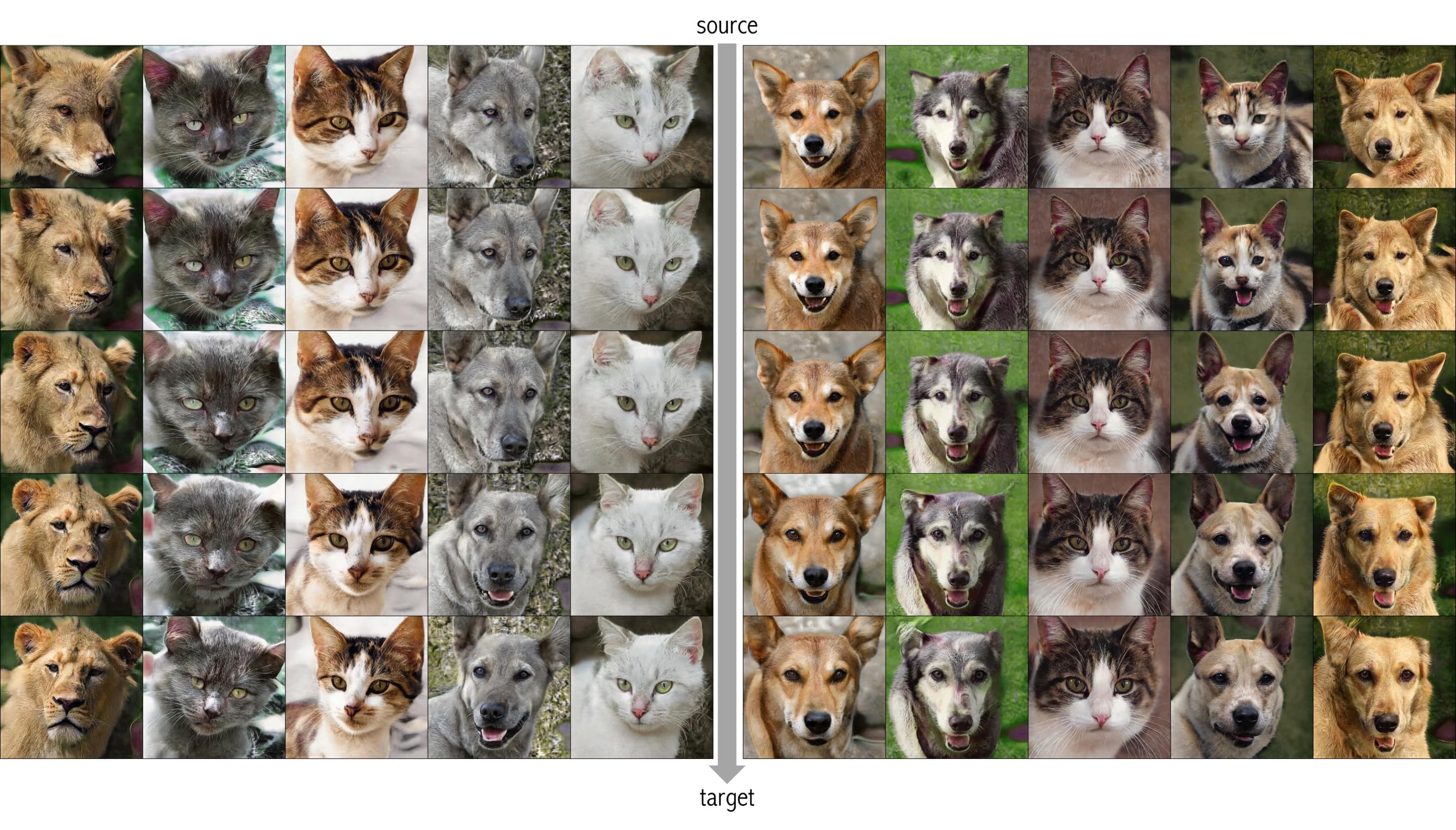}
\vspace*{-0.5cm}
\caption{Examples from interpolated content codes. All images are samples by varying the attention maps across all layers. All samples show that we can control content information independently from  the styles.}
\label{fig:f02}
\end{figure*}

\begin{figure*}[!h]
\centering
\includegraphics[width=1.0\linewidth]{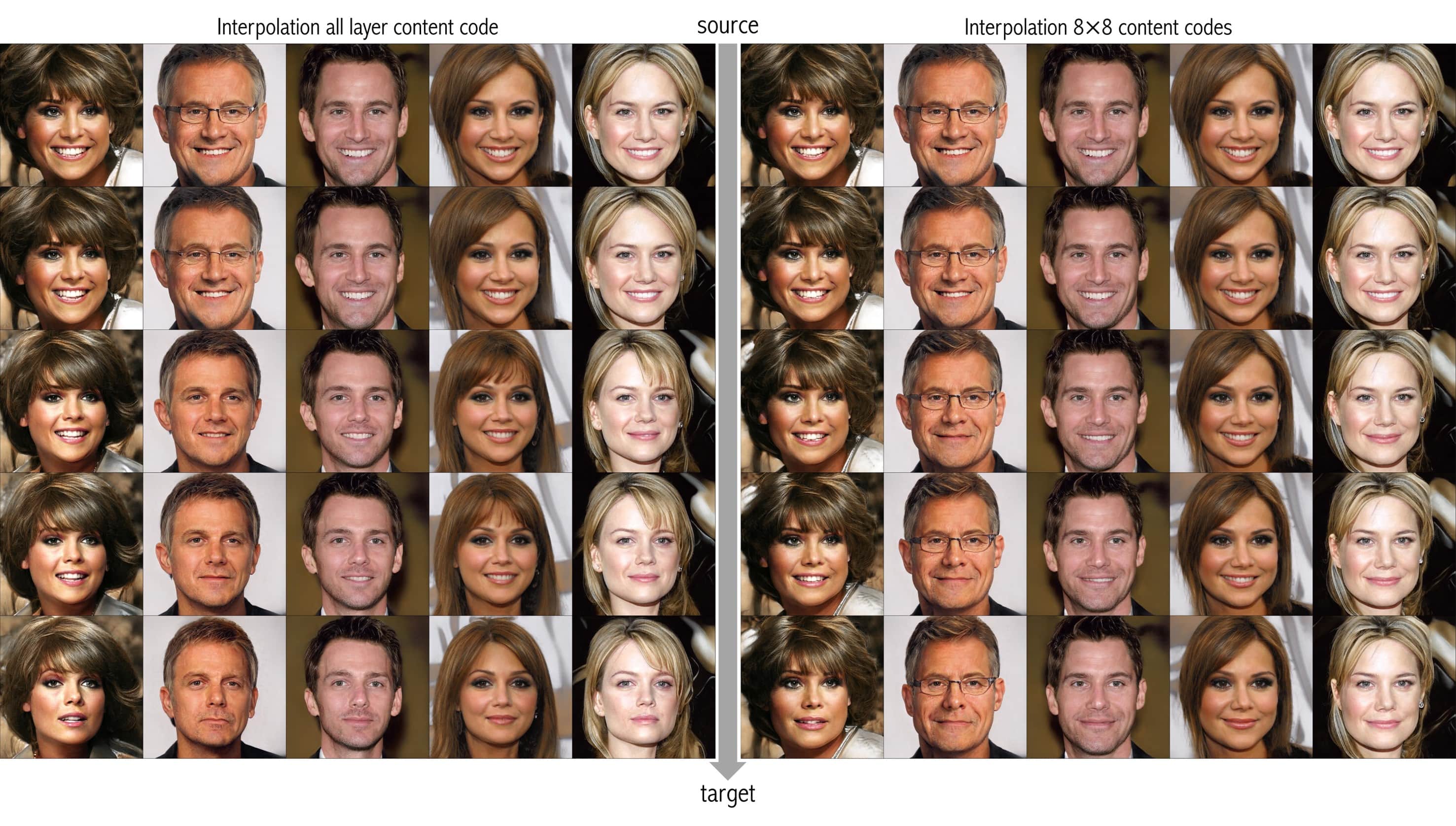}
\vspace*{-0.5cm}
\caption{Examples from interpolated content codes. The images at each row are sampled from the same content code. (Left) Samples by varying the content codes across entire layers. We can observe that the global attributes including rotation, facial expression, and identity changes gradually. (Right) Samples with changing 8$\times$8 resolution content codes. We can see the lower part of faces (i.e mouth) changes. }
\label{fig:c1}
\end{figure*}

\begin{figure*}[!h]
\centering
\includegraphics[width=1.0\linewidth]{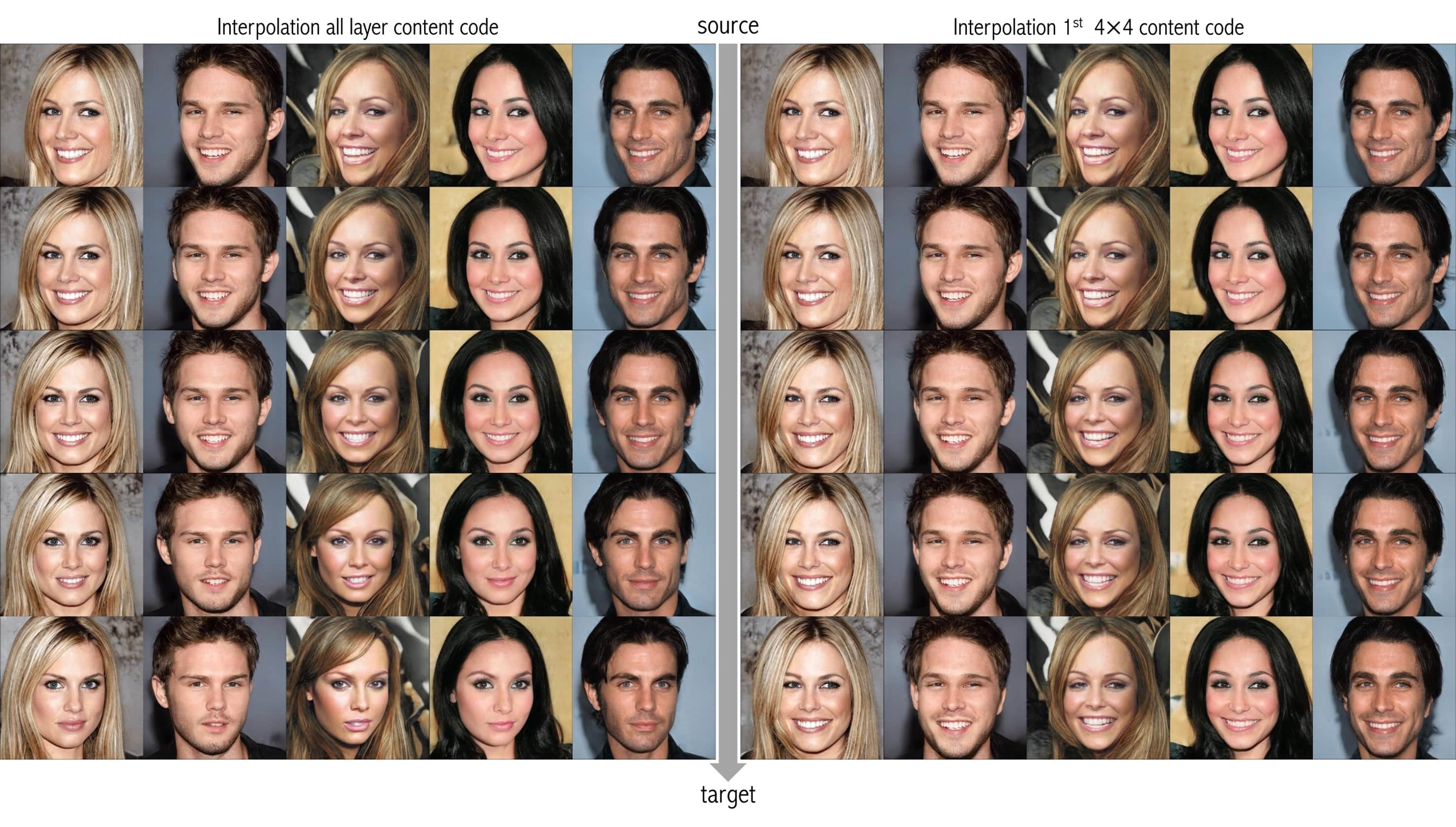}
\vspace*{-0.5cm}
\caption{Examples from interpolated content codes. The images of each row are sampled from same content code. (Left) samples with changing content code across all layers. (Right) Samples by varying  the first 4$\times$4 resolution content code. We can see the global geometry (i.e rotation) changes.}
\label{fig:c2}
\end{figure*}


\begin{figure*}[!h]
\centering
\includegraphics[width=1.0\linewidth]{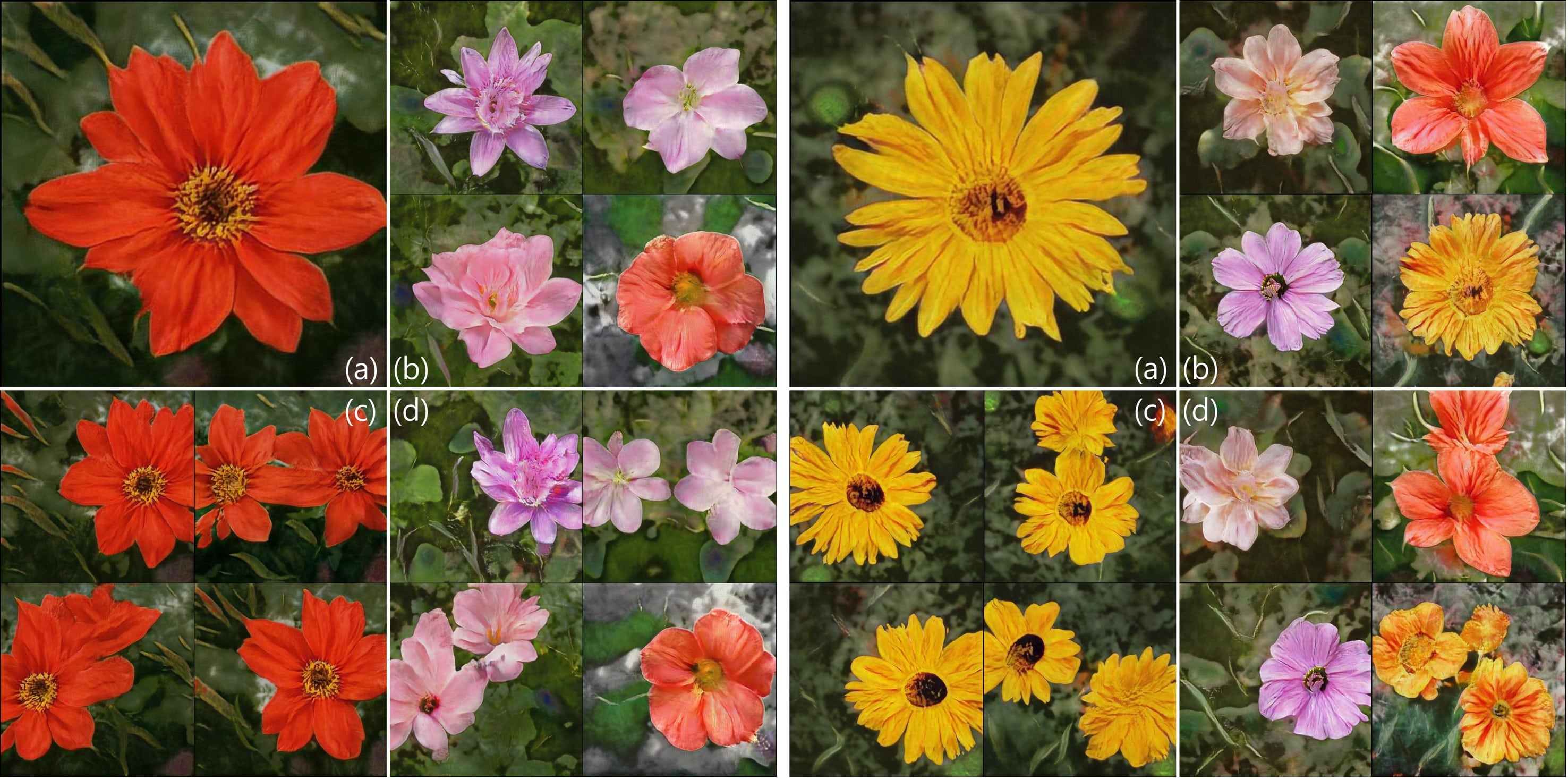}
\vspace*{-0.3cm}
\caption{Generated 512$\times$512 images by our method 
trained using flower data set. (a) A source image generated from arbitrary style and content code. (b) Samples with varying style codes and the fixed content code. We can observe the changes of the global style attributes including species, flower color and background. (c) Samples generated with varying content codes and the fixed style. We can observe that the content attributes including location, global shape, and the number of flowers change with different content codes. (d) Samples generated with both varying content and style codes.}
\label{fig:f1}
\end{figure*}
\vfill
\begin{figure*}[!h]
\centering
\includegraphics[width=1.0\linewidth]{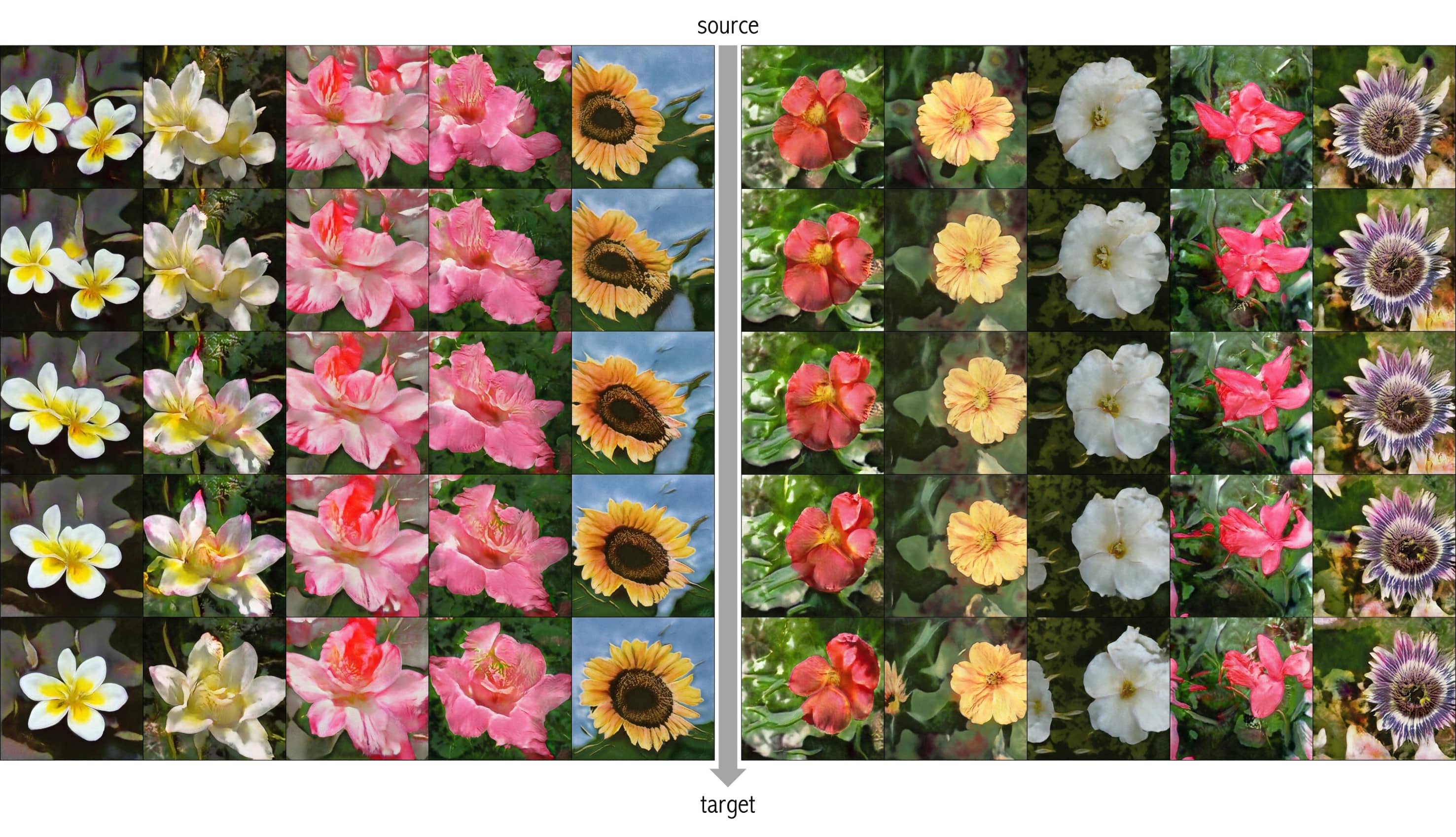}
\vspace*{-0.7cm}
\caption{Examples from interpolated content codes. The images of each row are sampled from the same content code. All images are samples with changing content codes across all layers.}
\label{fig:f2}
\end{figure*}

\begin{figure*}[!h]
\centering
\includegraphics[width=1.0\linewidth]{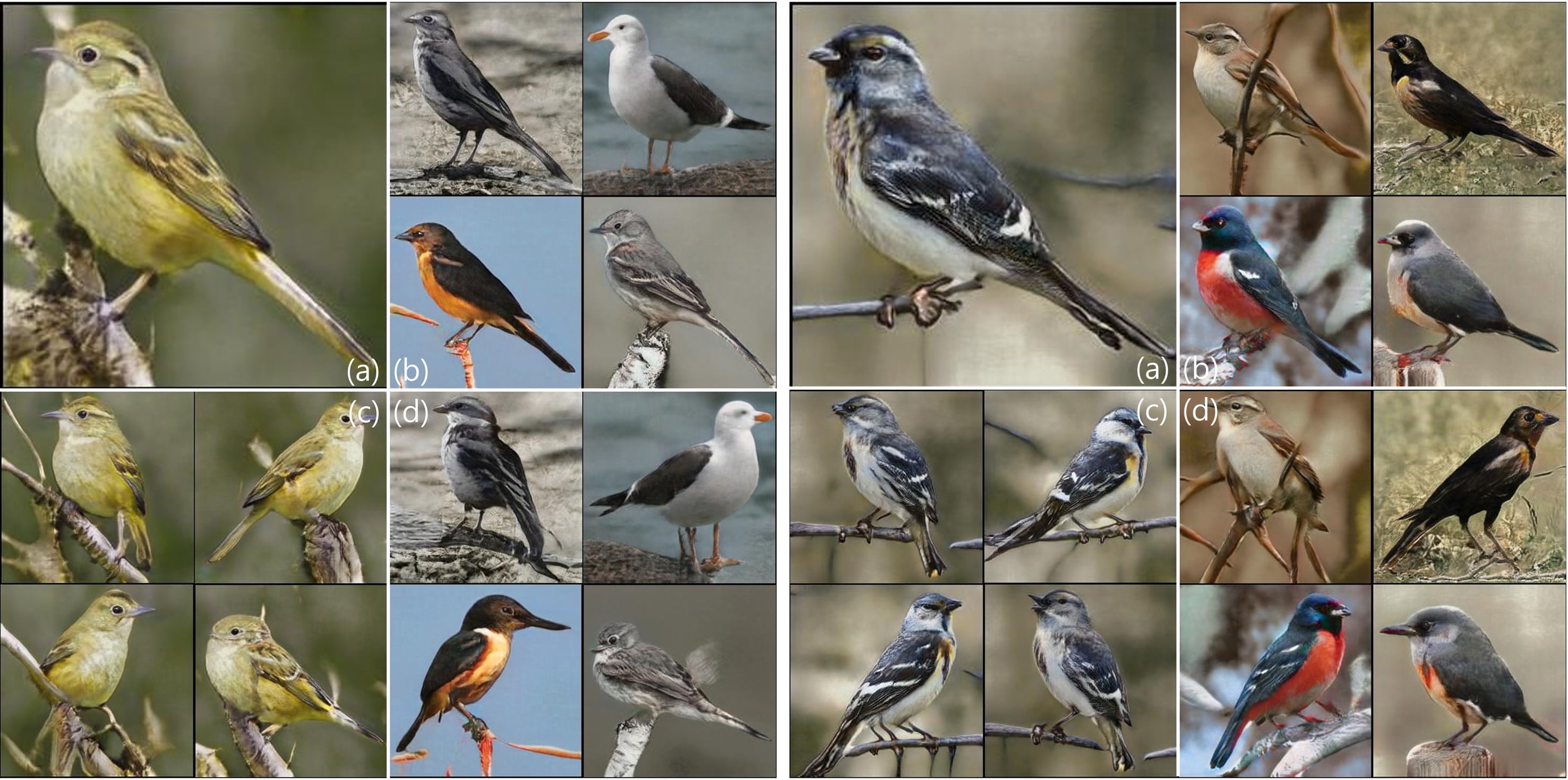}
\caption{Generated 256$\times$256 images by our method 
trained using bird data set. (a) A source image generated from arbitrary style and content code. (b) Samples with varying style codes and the fixed content code. We can observe the changes of  the global style attributes including species, feather colors and patterns. (c) Samples generated with varying content codes and the fixed style. We can observe that the content attributes including location, rotation and global shapes change with different content codes. (d) Samples generated with both varying content and style codes.}
\label{fig:c4}
\end{figure*}
\vfill
\begin{figure*}[!t]
\centering
\includegraphics[width=1.0\linewidth]{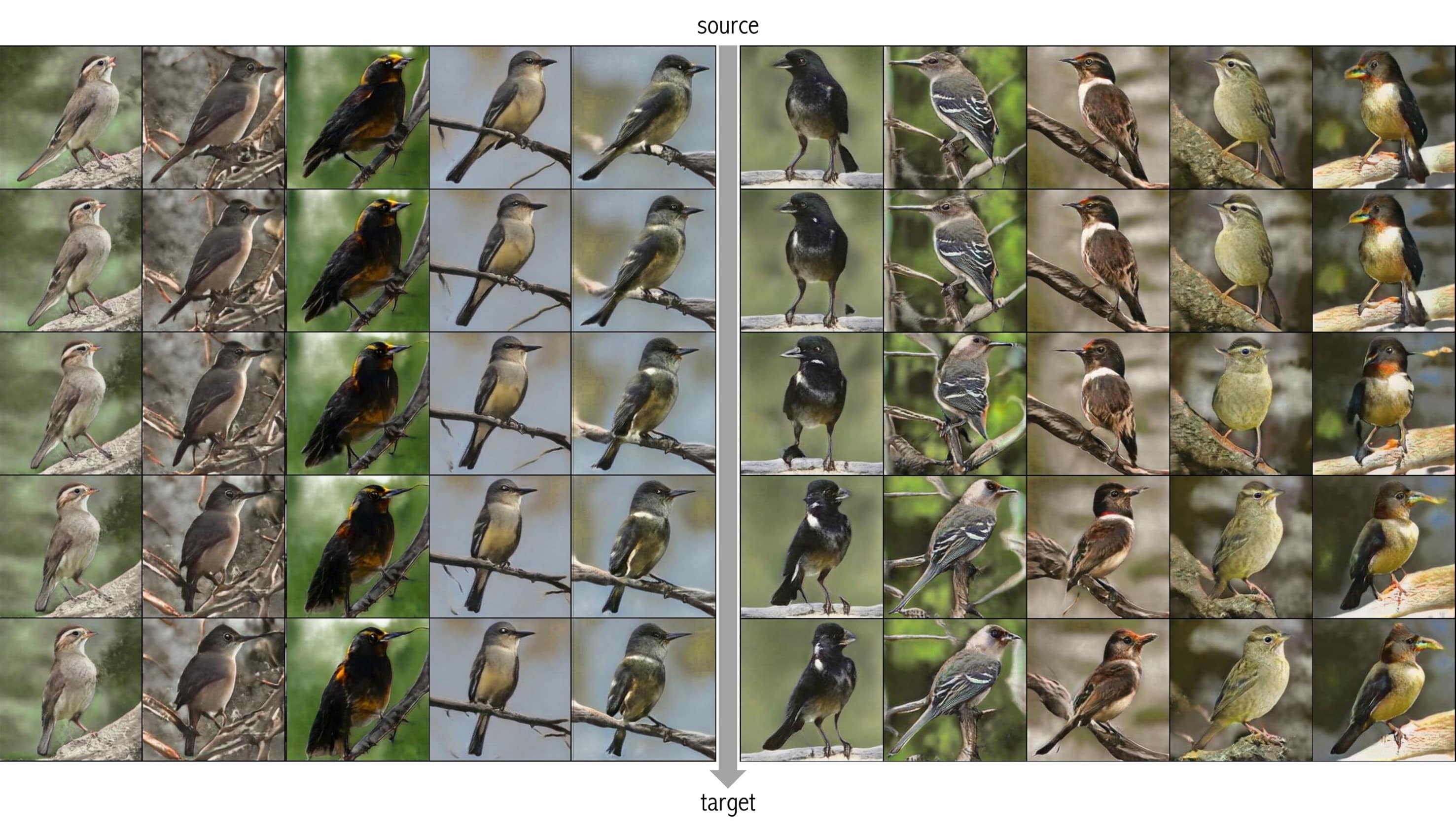}
\vspace*{-0.7cm}
\caption{Examples from interpolated content codes. The images of each row are sampled from the same content code. All images are samples with changing content codes across all layers.}
\label{fig:c5}
\end{figure*}

\begin{figure*}[!h]
\centering
\includegraphics[width=1.0\linewidth]{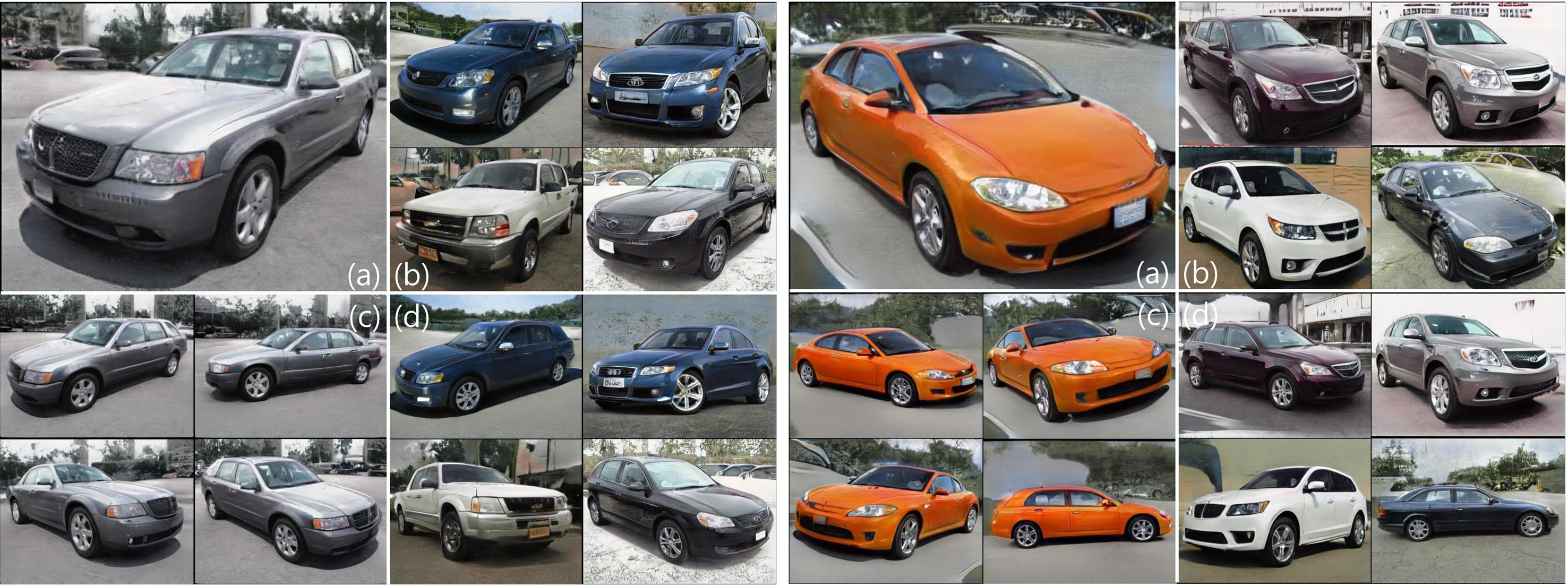}
\caption{Generated 384$\times$512 images by our method 
trained using car data set. (a) A source image generated from arbitrary style and content code. (b) Samples with varying style codes and the fixed content code. We can observe  the changes of the global style attributes including car type, colors and background. (c) Samples generated with varying content codes and the fixed style. We can observe that the content attributes including rotation and global shapes change with different content codes. (d) Samples generated with both varying content and style codes.}
\label{fig:c6}
\end{figure*}
\vfill
\begin{figure*}[!h]
\centering
\includegraphics[width=1.0\linewidth]{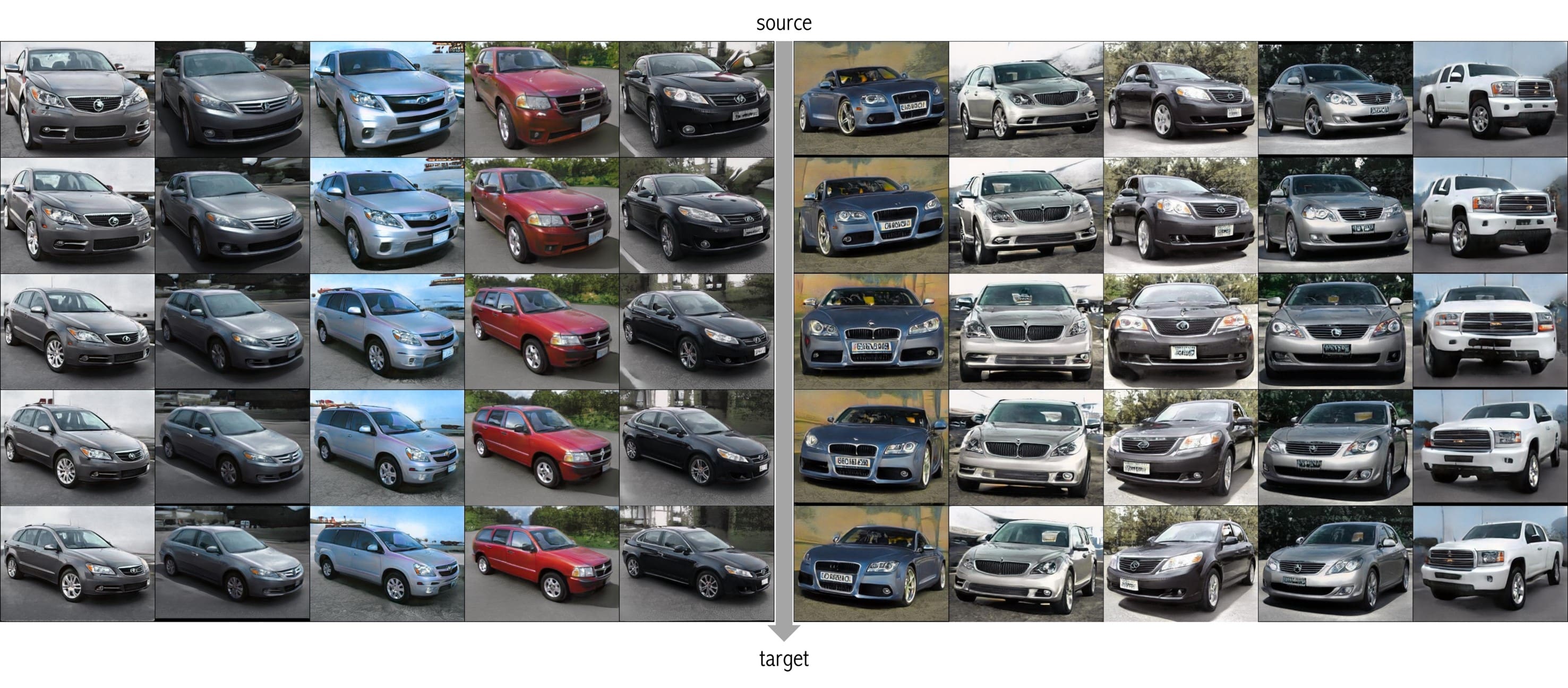}
\caption{Examples from interpolated content codes. The images of each row are sampled from the same content code. All images are samples with changing content codes across all layers.}
\label{fig:c7}
\end{figure*}
\vfill

\begin{figure*}[!h]
\centering
\includegraphics[width=0.8\linewidth]{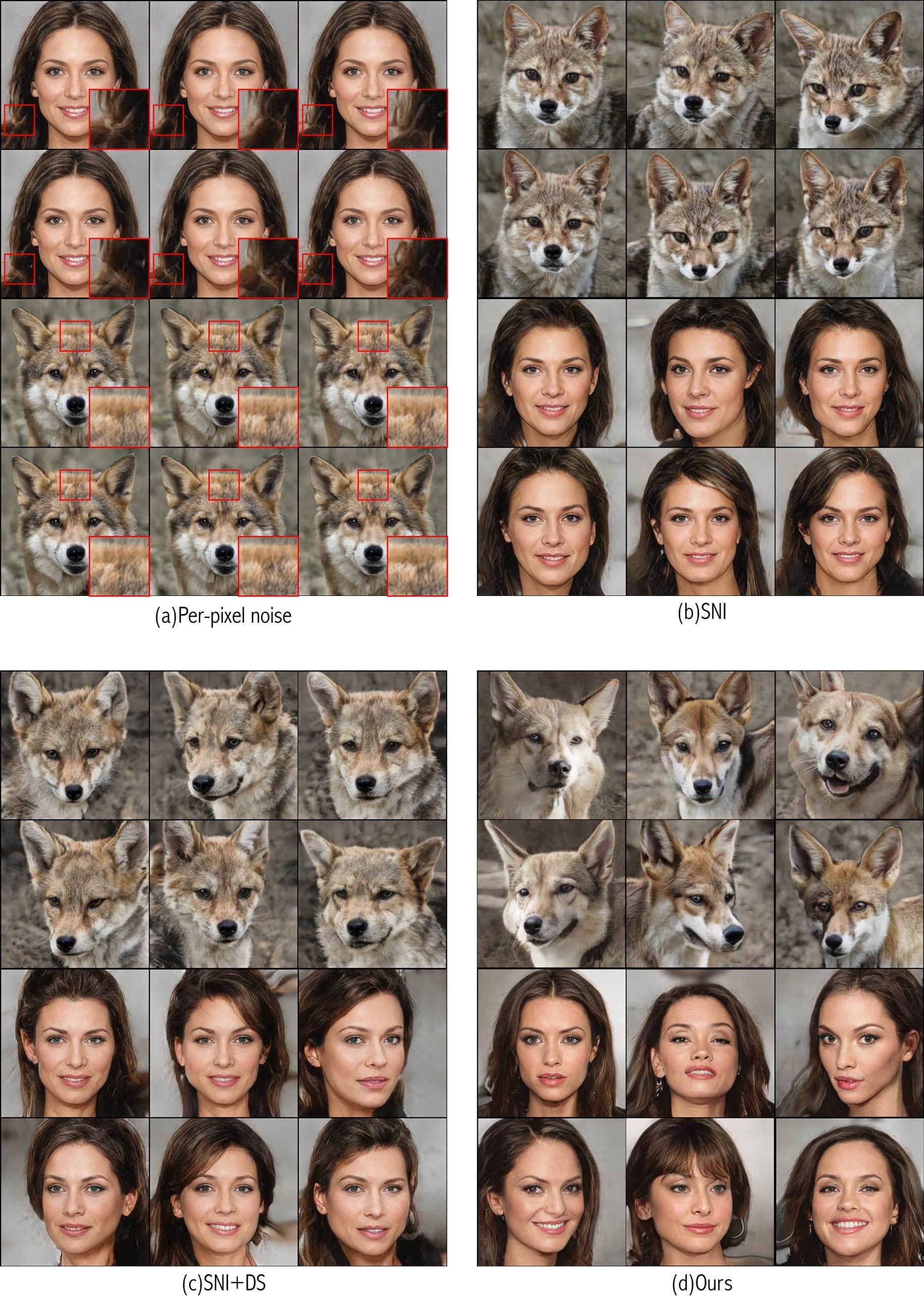}
\caption{Comparison results of various content controlling methods.  Sampled 256$\times$256 images from (a) StyleGAN with varying per-pixel noises, varying content codes of (b) SNI , (c) SNI trained with DS loss, and (d) our model.}
\label{fig:compare_sup}
\end{figure*}

\begin{figure*}[!h]
\centering
\includegraphics[width=1.0\linewidth]{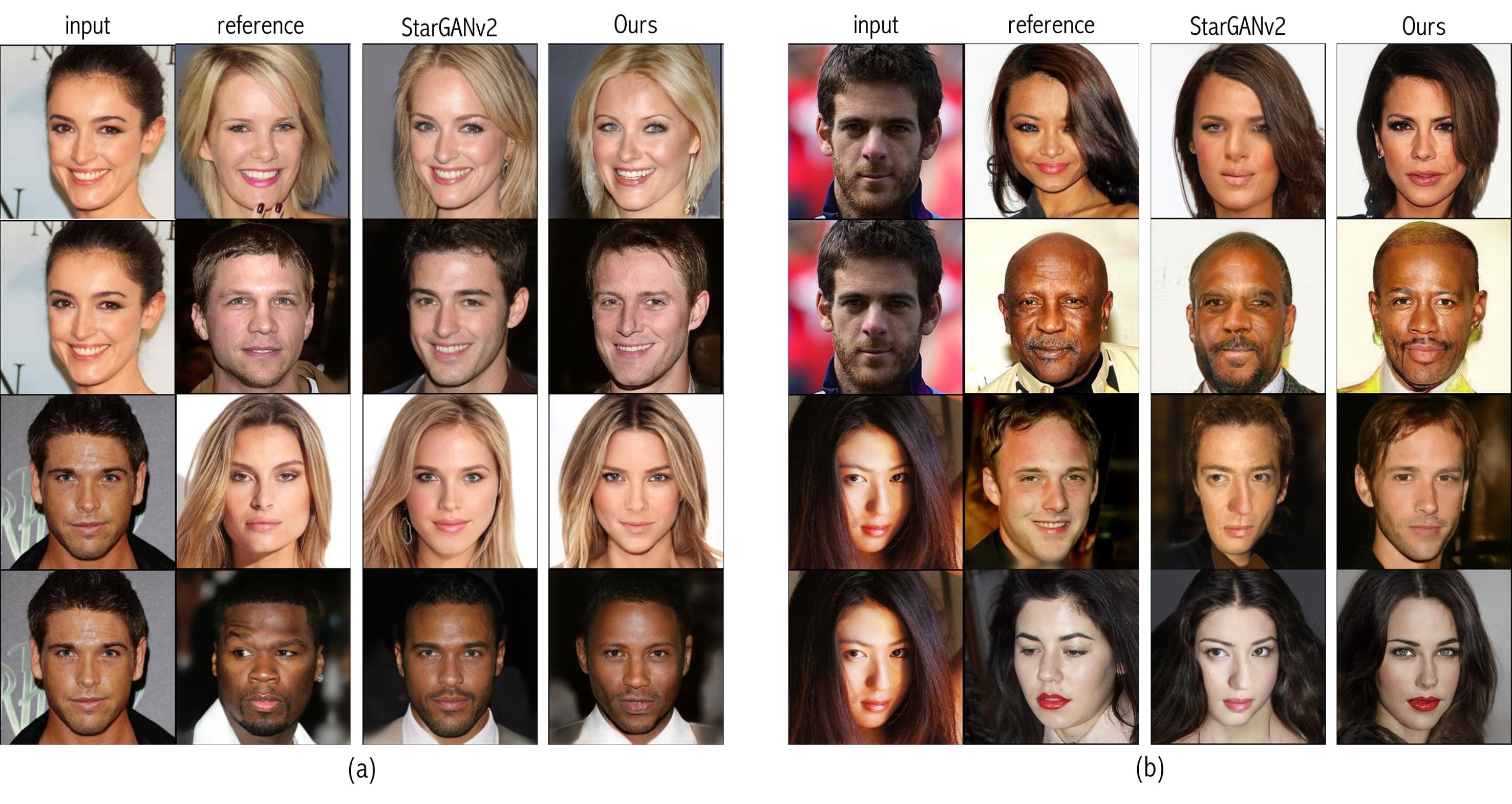}
\caption{Comparison results of reference-based synthesis. (a) Results from typical input images. Both StarGANv2 and ours can generate realistic images with the styles of reference images.  (b) Results from rare cases of input images. Our inversion model can successfully synthesize realistic images, whereas StarGANv2 fails.}

\label{fig:compare_reference}
\end{figure*}

\begin{figure*}[!h]
\centering
\includegraphics[width=.75\linewidth]{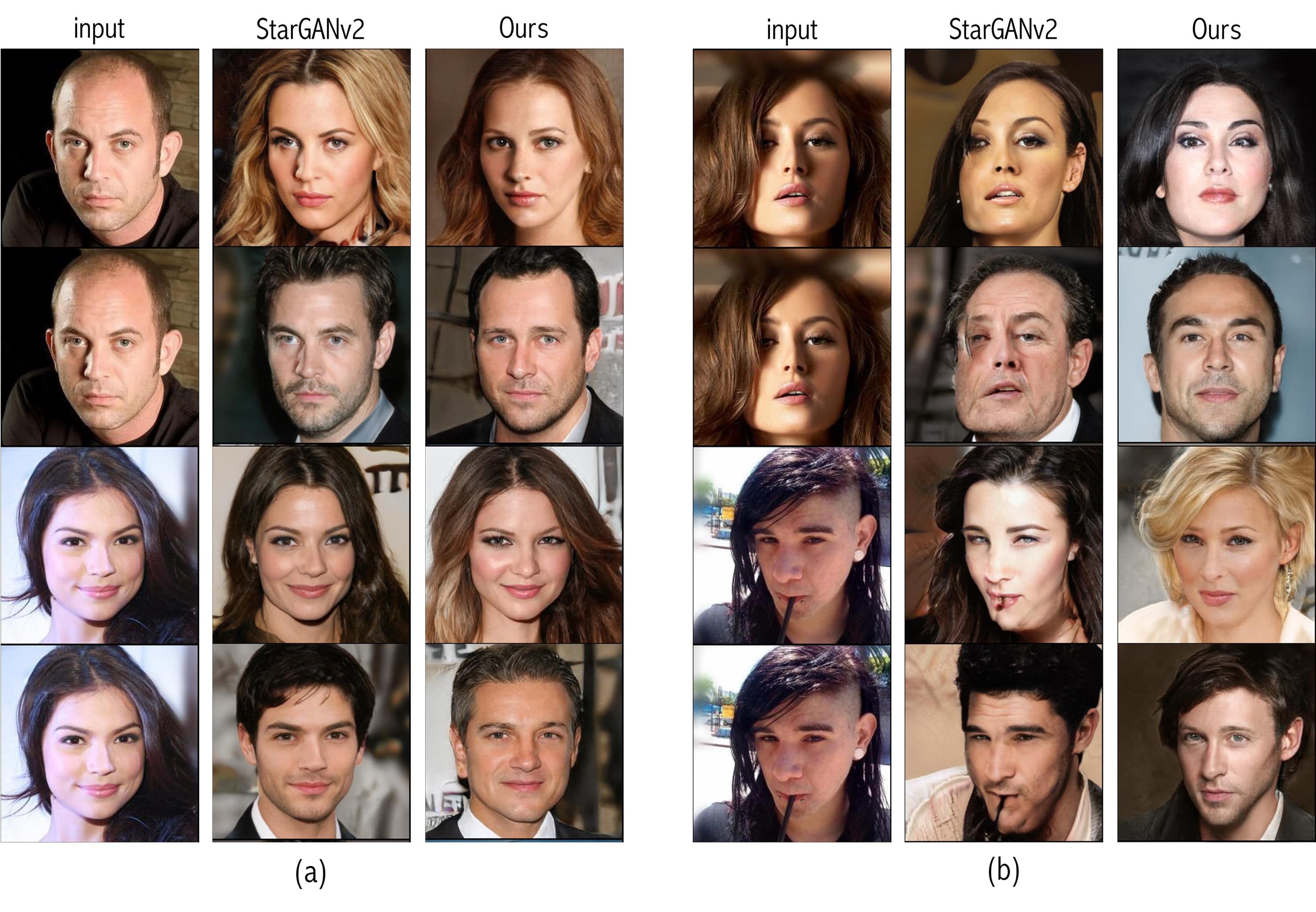}
\caption{Comparison results of latent-based synthesis. (a) Results from typical input images. Both StarGANv2 and Ours can generate realistic images with different styles. (b) Results from rare cases of input images. Our inversion model can successfully synthesize realistic images,
whereas StarGANv2 fails.}
\label{fig:compare_latent}
\end{figure*}

\begin{figure*}[!h]
\centering
\includegraphics[width=0.7\linewidth]{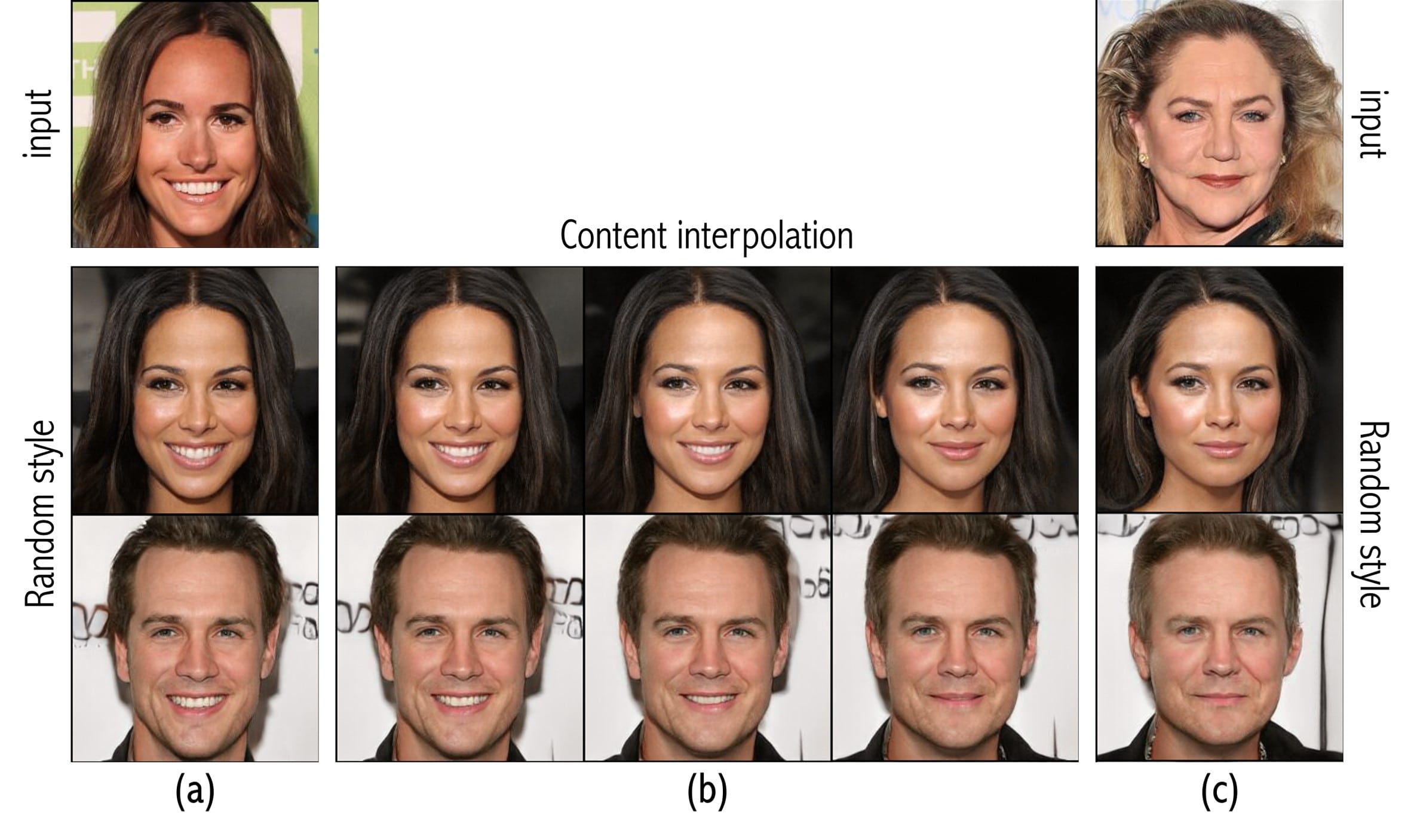}
\vspace{-0.5cm}
\caption{Results of content code interpolation from our model. (a,c) Similar to StarGANv2, our model can randomly change the style of the input images. (b) When interpolating the input content code between those of (a) and (c), we can obtain smooth content variation between two images. This variation is impossible in the existing image translation model such as StarGANv2.}
\label{fig:compare_intp}
\end{figure*}

\begin{figure*}[!h]
\centering
\includegraphics[width=0.9\linewidth]{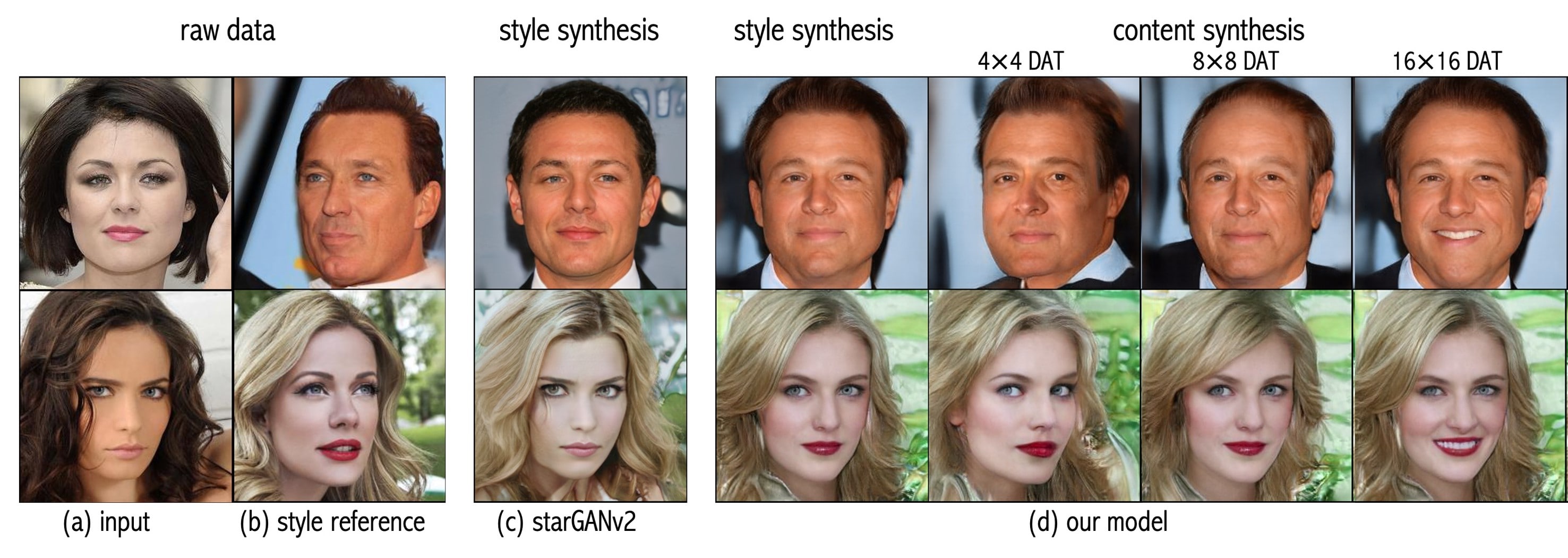}
\caption{Comparison between StarGANv2 and our model. (a) Input image and (b) style reference images.
Similar to  StarGANv2 in (c), our model can synthesize the style using the style reference as shown in the leftmost column of (d). 
Furthermore, by changing the content codes in a hierarchical manner through DAT layers,
 the corresponding content attributes are selectively synthesized as shown in from the second to the fourth columns in (d). This fine content control is not possible using StarGANv2.
}
\label{fig:compare_content}
\end{figure*}

\end{document}